\documentclass[a4paper,fleqn]{cas-sc}

\usepackage{bm}
\usepackage[numbers]{natbib}
\usepackage{url}
\usepackage{color}
\usepackage{graphicx}
\usepackage{algorithm}
\usepackage{algorithmic}

\newtheorem{theorem}{Theorem}
\newproof{pf}{Proof}

\def\tsc#1{\csdef{#1}{\textsc{\lowercase{#1}}\xspace}}
\tsc{WGM}
\tsc{QE}
\tsc{EP}
\tsc{PMS}
\tsc{BEC}
\tsc{DE}

\begin{document}
\let\WriteBookmarks\relax
\def\floatpagepagefraction{1}
\def\textpagefraction{.001}
\shorttitle{Variational Bayesian Weighted Complex Network Reconstruction}
\shortauthors{Xu et~al.}

\title [mode = title]{Variational Bayesian Weighted Complex Network Reconstruction}                      

\author[1]{Shuang Xu}[orcid= 0000-0003-3576-6914 ,style=chinese,bioid=1]
\cormark[1]
\ead{shuangxu@stu.xjtu.edu.cn}
\credit{Writing - original draft, Formal analysis, Funding acquisition, Methodology, Software, Visualization}
\address[1]{School of Mathematics and Statistics, Xi'an Jiaotong University, Xi'an 710049, China}

\author[1]{Chunxia Zhang}[style=chinese,bioid=2]
\credit{Data curation, Formal analysis, Funding acquisition, Writing - review \& editing}
	
\author[2,3,4]{Pei Wang}[style=chinese,bioid=3]
\credit{Conceptualization, Formal analysis, Funding acquisition, Writing - review \& editing}
\address[2]{School of Mathematics and Statistics, Henan University, Kaifeng 475004, China}
\address[3]{Bioinformatics Center, Henan University, Kaifeng 475004, China}
\address[4]{Institute of Applied Mathematics, Laboratory of Data Analysis Technology, Henan University, Kaifeng 475004, China}

\author[1]{Jiangshe Zhang}[style=chinese,bioid=4]
\credit{Funding acquisition, Project administration, Supervision,Writing - review \& editing}
\cormark[2]

\cortext[cor1]{Corresponding author}
\cortext[cor2]{Principal corresponding author}

\begin{abstract}
Complex network reconstruction is a hot topic in many fields. Currently, the most popular data-driven reconstruction framework is based on lasso. However, it is found that, in the presence of noise, lasso loses efficiency for weighted networks. This paper builds a new framework to cope with this problem. The key idea is to employ a series of linear regression problems to model the relationship between network nodes, and then to use an efficient variational Bayesian algorithm to infer the unknown coefficients. The numerical experiments conducted on both synthetic and real data demonstrate that the new method outperforms lasso with regard to both reconstruction accuracy and running speed.
\end{abstract}


\maketitle

\section{Introduction}\label{sec:1}
The networked-systems are ubiquitous in many fields, including social-tech science \cite{Social1,Epidemic2}, bioinformatics \cite{Biology2,Biology,Biology0,Biology3},
epidemic dynamics \cite{Epidemic0,Epidemic3,Epidemic} and power grid \cite{PowerGrid2,PowerGrid}. As is often the case, it is unable to observe the topology of a network, while data generated by this network is available. Therefore, in interdisciplinary science, one of the most important but challenging problems is to reconstruct the complex network from observed data or time series \cite{recon_review}.

Suppose that a complex network consists of $N$ nodes, and we are given the time series of the states for the $N$ nodes. A decade ago, the main network reconstruction technique was causality analysis (CA) \cite{GrangerRecon,GrangeCausality,PartialGrangerCausality,PPGrangerCausality,PiecewiseGrangerCausality} which infers the causal influence between two variables via a pair of linear regression models. With the rapid development of variable selection and feature learning \cite{BayesLasso,lasso,VS}, CA has been replaced with lasso \cite{lasso_ori} (a.k.a. compressive sensing). Wang et al. \cite{CS_GT} made the first attempt to apply lasso to reconstructing networks with game-theoretic dynamics. Thereafter, Wang's group applied lasso to networks with epidemic \cite{CS_SIS}, geospatial \cite{CS_geo} and electrical power \cite{CS_CN} data. They also studied the application to related problems (e.g., the prediction of catastrophes in nonlinear dynamical systems \cite{CS_Catastrophes}). Recently, Wu's group has extended this framework to multilayer networks \cite{CS_multi} and the networks with time-varying nodal parameters\cite{CS_tv}. To some extent, lasso is a ``black-box'' tool for the complex network reconstruction task. Recently, an increasing number of researchers start to develop alternative techniques to lasso. For example, Ma et al. and Xiang
et al. cast the problem into a statistical inference issue \cite{Recon_EM,Recon_EM2}.

Essentially, lasso \cite{lasso_ori} solves an $L_1$-norm penalized least squares problem, i.e.,  $\min_{\bm{w}\in R^N}\|\bm{y}-\bm{Xw}\|^2_2+\lambda_{\rm lasso}||\bm{w}||_1$. Owing to the property of the $L_1$-norm penalty \cite{CS}, the solution of $\bm{w}$ is sparse, that is, many entries of $\bm{w}$ are zero. The parameter $\lambda_{\rm lasso}>0$ controls the sparsity, and larger $\lambda_{\rm lasso}$ makes $\bm{w}$ sparser. In the context of (unweighted) network reconstruction, $\bm{y}$ and $\bm{X}$ are observed data, and the element of regression coefficient vector $\bm{w}$ indicates whether a pair of nodes are connected or not. In other words, $\bm{w}$ corresponds to a column of the adjacent matrix $\bm{A}$ of a network. Therefore, the network topology can be recovered column-by-column via applying lasso $N$ times.

\begin{figure}
	\centering
	\includegraphics[scale=.75]{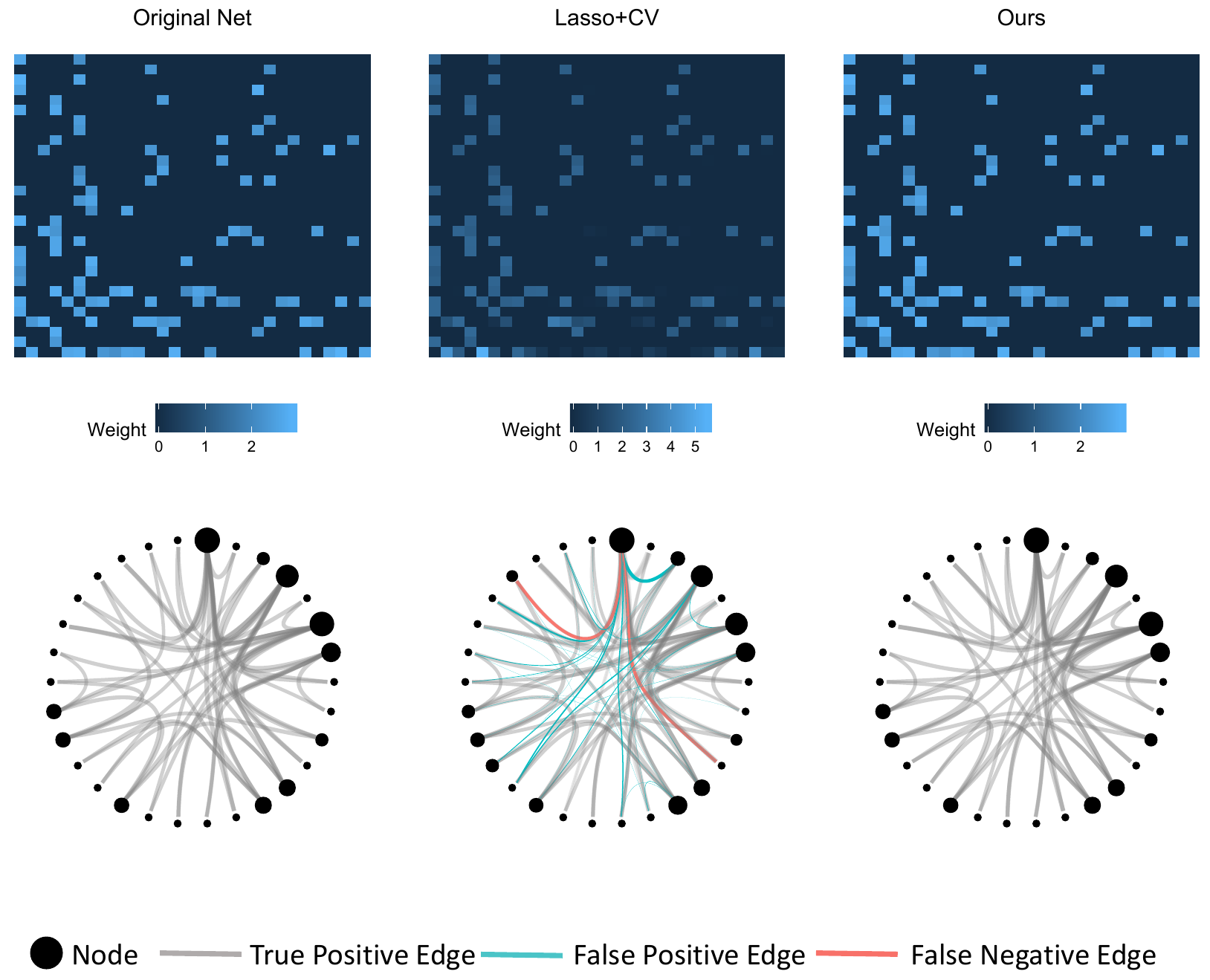}
	\caption{Top: The heatmap of networks. Note that the $(i,j)^{\rm th}$ cell of heatmap corresponds to $(i,j)^{\rm th}$ entity of weighted matrix $\bm{W}$. Bottom: The visualization of the original network as well as the reconstructed ones by our method and lasso with CV, respectively. The size of each node and edge is proportional to its degree and weight, respectively. It is shown that our method correctly identifies all the edges, while lasso with CV fails to detect 2 existing edges (i.e., the false negative edges colored by red) and identifies 36 fake edges (i.e., the false positive edges colored by blue). The error of connection strength of lasso and our method is 0.297 and 0.006, respectively. The running time of lasso and our method is 0.851 and 0.016 second, respectively. \textbf{Simulation details}: the original network $\bm{W}$ is a BA network of 30 nodes; the number of data points is $M=30$ and the scale of noise is $\sigma=0.05$. }
	\label{fig:fig1a}
\end{figure}

Unfortunately, the current researchers pay little attention to the weighted network reconstruction task. In the cases of weighted networks, besides determining whether each pair of nodes are connected or not, it is also required to estimate the connection strength. To meet the demands, lasso based framework has to carefully select $\lambda_{\rm lasso}$. In general, a good $\lambda_{\rm lasso}$ can be obtained by cross validation (CV) technique \cite{CrossValidation}. But it is a time-consuming strategy. Additionally, lasso in theory cannot precisely estimate the connection strength owing to lasso's \textit{biased estimation property} \cite{lasso_ori}. A toy example shown in Fig. \ref{fig:fig1a} illustrates the drawback of lasso. To the best of our knowledge, there are only two references related to weighted network reconstruction, but they focus on small-scale networks \cite{CS_weight,CS_weight2} (the number of nodes is less than 10). 

By putting the network reconstruction task into the hierarchical Bayesian modeling framework, we in this paper propose an elegant solution to weighted complex network reconstruction, especially for large-scale networks. This work contains the following threefold contributions. 

(i) Firstly, to overcome the shortcoming of lasso, we consider a special regression problem with multiple regression coefficients, that is, $\min_{\bm{w}, \bm{a}} \|\bm{y}-\bm{X}(\bm{a}\odot\bm{w})\|^2$, where $\odot$ denotes the element-wise product, $\bm{w}\in R^N$ plays the same role as the traditional regression coefficient vector, and $\bm{a}\in \{0,1\}^N$ indicates variables are active or not. In the context of complex networks, the real-valued $\bm{w}$ is used to estimate the connection strength, while the binary-valued $\bm{a}$ represents whether nodes are connected or not. In later discussions, it will be found that this formulation is very suitable for handling large-scale network reconstruction problems. 

(ii) Secondly, a full hierarchical Bayesian inference technique is used to estimate the unknown parameters ${\bm{w}}$ and ${\bm{a}}$. By the virtue of hierarchical Bayesian inference, our model overcomes the shortcoming of lasso, i.e., tuning hyper-parameters. To speed up the inference, a variational Bayesian technique rather than Markov chain Monte Carlo (MCMC) sampling is employed to approximate the posterior distributions of unknown variables \cite{VB,VB2,VB3}. To the best of our knowledge, this is the first attempt to utilize the variational Bayesian technique to reconstruct networks. 

(iii) The experiments conducted on simulated networks (scale-free and small-world networks with electrical current transportation (ECT) and communication dynamics) and a stock network show that our framework significantly outperforms lasso.

The rest of this paper is organized as follows. In Section 2, we provide the definition of the network reconstruction task. In Section 3, we formulate the new framework and show how to infer the network topology with it. Then, some experiments are conducted in Section 4. Finally, the paper ends with conclusions and future works.

\begin{table}
	\centering
	\caption{ The list of notations in the paper.  }	\label{tab:1}
	\begin{tabular}{c|c}
		\hline
		Notation& Description \\	\hline
		$ \boldsymbol{A},\boldsymbol{W} $& The adjacency and weighted matrix of a network \\ 	\hline
		$ \hat{\boldsymbol{A}},\hat{\boldsymbol{W}} $& The reconstructed network \\ 	\hline
		$ \boldsymbol{y} $&  Response vector\\ 	\hline
		$ \boldsymbol{X} $& Design matrix \\ 	\hline
		$ \boldsymbol{w} $& Regression coefficient vector \\	\hline
		$ \boldsymbol{a} $& Binary regression coefficient vector \\ 	\hline
		$N,M$& The number of nodes and observed data, respectively  \\ 	\hline
		$\epsilon,\sigma^2$& Noise item and its variance \\  \hline
		$c_0,d_0,g_0,h_0$& Parameters in Gamma distribution \\  	\hline
		$e_0,f_0$& Parameters in Beta distribution \\ 	\hline
		$\rho$& A parameter in Bernoulli distribution \\  \hline
		$\lambda_j$& A parameter in the distribution of $w_j$ \\  \hline
	\end{tabular}
\end{table}

\section{Problem Statement}\label{sec:related_work}

To facilitate later discussions, Table \ref{tab:1} lists the main notations and symbols used in the paper. By following the common practice of many references \cite{CS_CN,CS_weight,CS_multi,CS_SIS,CS_geo,CS_GT,CS_Catastrophes,CS_tv,CS_weight2}, we consider a weighted complex network without loops. Suppose that its weighted matrix is $\bm{W}=(w_{ij})$, where $w_{ij}$ represents the connection strength from node $i$ to $j$. We set $w_{ij}=0$ if nodes $i$ and $j$ are not connected. Its adjacency matrix is $\bm{A}=(a_{ij})$, where $a_{ij}=1$ if $w_{ij}\neq0$ and 0 otherwise. Moreover, we assume that the nodes are governed by a specific dynamics. As defined in the reference \cite{recon_review}, the network reconstruction task is to infer the network topology according to observed data or time series from nodal dynamics. 

Here, to facilitate illustration, we take the ECT in a power network consisting of resistors as a special example to introduce the network reconstruction task. The resistance of a resistor between nodes $i$ and $j$ is denoted by $r_{ij}$, where $r_{ij}=\infty$ if they are not connected. Based on the Kirchhoff's law, there is
\begin{equation}\label{eq1}
\sum_{j=1}^{N} \frac{a_{ij}}{r_{ij}}(V_i-V_j)=I_i \quad (i=1,2,\cdots,N),
\end{equation}
where $V_i$ and $I_i$ denote the voltage and the electrical current of node $i$, respectively. Generally speaking, in the real world, we are able to observe the nodes' voltage and electrical current, while the network structure is invisible. If we record the voltage and electrical current at different time points, the reconstruction task for this power network is to infer the network topology (including $a_{ij}$ and $r_{ij}$) based on the observed data. 

\section{Bayesian complex network reconstruction} \label{sec2}
\subsection{Reconstructing network by regression}
In this subsection, we show that the network reconstruction task can be accomplished by solving a series of regression problems. In what follows, we define $x_{j}^{(i)}=V_i-V_j$ and $w_{ij}=1/r_{ij}$. At the same time, we use $y^{(i)}$ to refer to $I_i$. Since the network simulates a dynamic system, it is reasonable to assume that data are collected at $M$ different time points. To facilitate description, let $\{t_m\}_{m=1}^{M}$ denote the time index set. With the above assumptions, according to Eq. (\ref{eq1}), the variables should satisfy
\begin{equation} \label{eq2}
\left[ \begin{matrix} y^{(i)}_{t_1} \\ y^{(i)}_{t_2} \\ \vdots \\ y^{(i)}_{t_M}	\end{matrix} \right]  = 	\left[ \begin{matrix} x^{(i)}_{1,t_1} & x^{(i)}_{2,t_1} & \cdots & x^{(i)}_{N,t_1} \\ x^{(i)}_{1,t_2} & x^{(i)}_{2,t_2} & \cdots & x^{(i)}_{N,t_2} \\ \vdots & \vdots & & \vdots \\ x^{(i)}_{1,t_M} & x^{(i)}_{2,t_M} & \cdots & x^{(i)}_{N,t_M}
\end{matrix}\right] \left[ \begin{matrix} a_{i1}w_{i1} \\ a_{i2}w_{i2} \\ \vdots \\ a_{iN}w_{iN}	\end{matrix} \right].
\end{equation}
It is noteworthy that the item $y_{t_m}^{(i)}=I_{i,t_{m}}$ represents the electrical current of node $i$ and $x_{j,t_m}^{(i)}=V_{i,t_{m}}-V_{j,t_{m}}$ denotes the difference of voltage between nodes $i$ and $j$ at time $t_{m}$. In the matrix form, there is
\begin{equation} \label{eq3}
\bm{y}^{(i)} = \bm{X}^{(i)}\mathbb{D}(\bm{a}^{(i)})\bm{w}^{(i)},
\end{equation}
where $\mathbb{D}(\bm{a}^{(i)})$ is a diagonal matrix whose main diagonal is $\bm{a}^{(i)}$, and $\bm{a}^{(i) }=\left( a_{i1},\cdots,a_{iN}\right)^{\rm T}$ is the $i^{\rm th}$ row of matrix $\bm{A}$. As stated above, the nodal states are observable. In other words, $\bm{y}^{(i)}$ and $\bm{X}^{(i)}$ in Eq. (\ref{eq3}) are observed data; $\bm{a}^{(i)}$ and $\bm{w}^{(i)}$ are unknown items. The network's weighted matrix can recovered column-by-column via solving Eq. (\ref{eq3}) with $i=1,\cdots,N$. In this manner, the network reconstruction can be cast into a series of regression problems. 

\subsection{Model formulation}
Since the network reconstruction actually corresponds to solving some special regression problems, in this subsection we cast it into a new framework based on Bayesian statistics.

In the following discussions, we reformulate the estimation problem in Eq. (\ref{eq3}) as a linear regression model via
\begin{equation}
\label{eq:new_model}
\bm{y}=\bm{X}\mathbb{D}(\bm{a})\bm{w}+\bm{\epsilon},
\end{equation}
where $\bm{\epsilon}$ denotes the noise item. The response vector $\bm{y}\in R^M$ and the design matrix $\bm{X}\in R^{M\times N}$ are observable, while the binary coefficient vector $\bm{a}$ and the continuous coefficient vector $\bm{w}$ need to be estimated. Here $\mathbb{D}(\bm{a})$ denotes a diagonal matrix whose main diagonal is the vector $\bm{a}$. We emphasize that this framework can handle weighted networks with many kinds of dynamics \cite{recon_review}, including communication dynamics \cite{CS_CN}, evolutionary game dynamics \cite{CS_GT}, epidemics \cite{CS_SIS}, synchronization dynamics \cite{CS_multi}, to name but a few. Besides ECT, we introduce how to apply our framework to communication dynamics in \textbf{Appendix B}. 

In general, it can be assumed that $\epsilon_{t_m}$ are independently and identically distributed (i.i.d.) as Gaussian, namely, $\epsilon_{t_m}\mid \sigma \sim\mathcal{N}(\epsilon_{t_m}\mid 0,\sigma^2), m=1,\cdots,M$, where $\sigma^2$ denotes the variance. Note that the popular reconstruction method lasso also falls into this category. The significant advantage of lasso over the simple least-squares method lies in that it can provide a sparse solution. However, its performance highly depends on the tuning of its parameter $\lambda_{\rm lasso}$. In contrast, Bayesian methods can provide satisfactory estimates for $\bm{w}$ and $\bm{a}$ while avoiding the tedious parameter adjustment. The core idea of Bayesian methods is to impose a prior distribution on each unknown variable (i.e., $\bm{w}$ and $\bm{a}$), then MCMC sampling or a variational technique is employed to approximate the posterior distribution according to the famous Bayes theorem. There are hyper-parameters in the prior distribution sometimes, and some proper hyper-prior distributions can be hypothesized. In what follows, we will adopt a full hierarchical Bayesian inference process to infer our interested items in Eq. (\ref{eq:new_model}).

\begin{figure}[pos=t]
	\centering
	\includegraphics[width=0.7\linewidth]{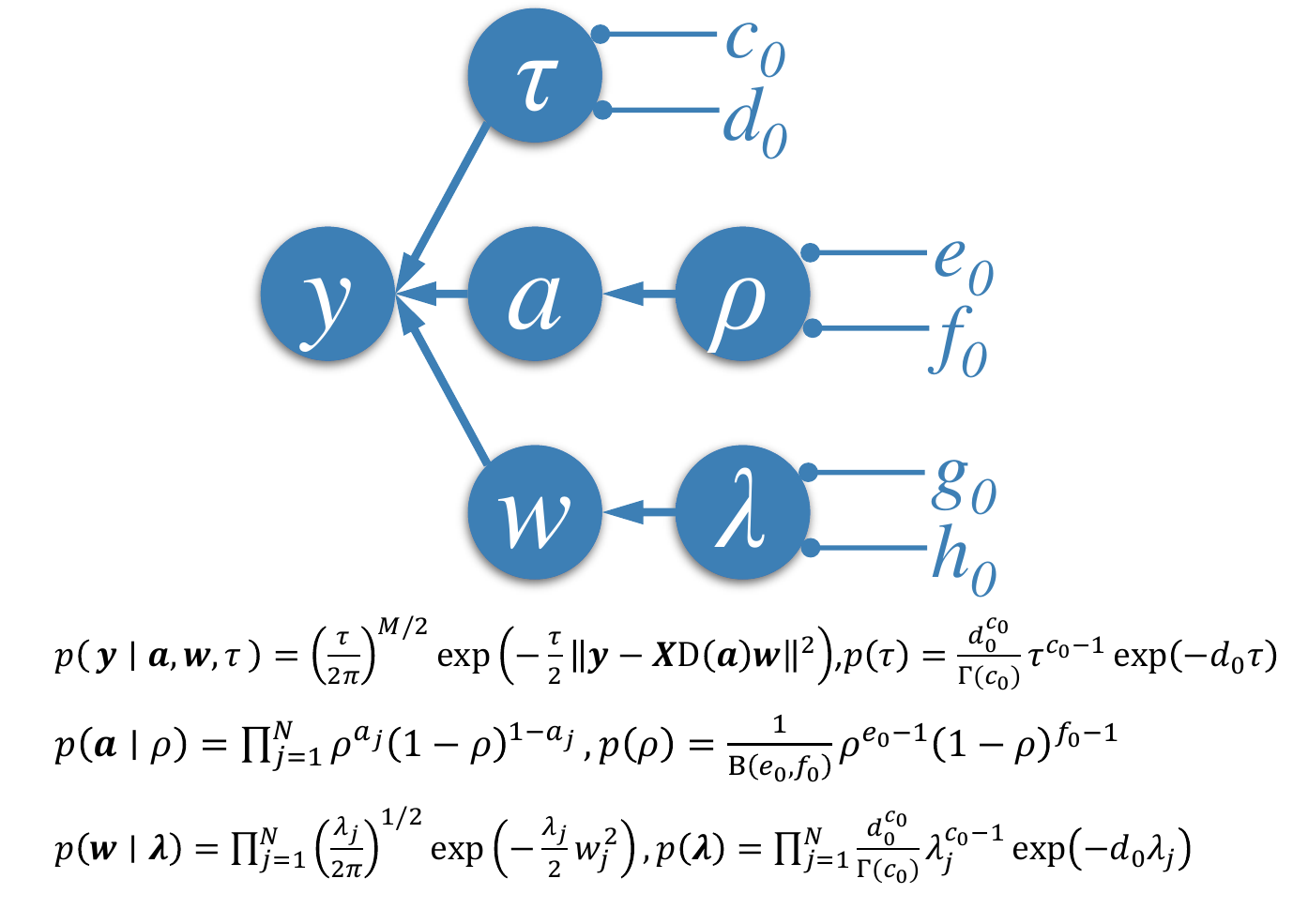}
	\caption{The graphical representation of the Bayesian complex network reconstruction model. Here $c_0,d_0,e_0,f_0,g_0,h_0$ are hyper-parameters. Note that $\Gamma(\cdot)$ and ${\rm B}(\cdot,\cdot)$ denote the gamma and beta functions, respectively.}
	\label{fig:fig1}
\end{figure}

To facilitate illustration, we let $\tau^{-1}\equiv\sigma^2$ and $\tau$ is often called precision. Thereafter, the conditional distribution of $\bm{y}$ is given by
\begin{equation}
\bm{y} \mid \bm{a}, \bm{w}, \tau \sim \mathcal{N}\left( \bm{y}\mid\bm{X}\mathbb{D}(\bm{a})\bm{w}, \tau^{-1}\bm{I}_{M}\right) ,
\end{equation}
where $\bm{I}_{M}$ denotes an identity matrix of order $M$. For each regression coefficient (connection strength) $w_j$, we place the following Gaussian prior
\begin{equation}
w_j \sim \mathcal{N}(w_{j}\mid 0,\lambda_{j}^{-1}), \quad j=1,2,\cdots,N,
\end{equation}
where $\lambda_{j}^{-1}$ is the variance. Since each entry $a_j$ of the coefficient vector $\bm{a}$ takes binary value $\{0,1\}$, it is natural to consider a Bernoulli prior for $a_{j}$, that is,
\begin{equation}
a_{j}\mid \rho \sim {\rm Bernoulli}\left( a_{j}\mid \rho\right) , \quad j=1,\cdots, N,
\end{equation}
where $\rho$ denotes the probability of $a_{j}$ taking value 1. At last, to make a full Bayesian inference, we impose conjugate priors on the parameters $\lambda_{j}$, $\tau$ and $\rho$, namely,
$
\tau \sim {\rm Gamma}(\tau\mid c_0,d_0), 
\rho \sim {\rm Beta}(\rho\mid e_0,f_0),
\lambda_{j} \sim {\rm Gamma} (\lambda_{j}\mid g_0, h_0),
$
where $c_0,d_0,e_0,f_0,g_0,h_0$ are hyper-parameters. To facilitate the understanding of the Bayesian framework, Fig. \ref{fig:fig1} shows the hierarchical Bayesian graph and the probability density functions of related random variables.

Subsequently, it is able to write the joint distribution of all variables, viz.,
\begin{equation}\label{eq:joint_distribution}
p\left( \bm{y},\bm{a},\bm{w},\bm{\lambda},\tau,\rho\right) 
=	 p\left( \bm{y}\mid \bm{a},\bm{w},\tau\right)p(\tau) \left[\prod_{j=1}^{N}p(a_{j}\mid\rho)p(w_j\mid\lambda_{j})p(\lambda_{j}) \right] p(\rho).
\end{equation}
Our aim is to obtain the posterior distribution of $\bm{a},\bm{w},\bm{\lambda},\tau$ and $\rho$. Based on the Bayes theorem, we have
\begin{equation}
p\left( \bm{a},\bm{w},\bm{\lambda},\tau,\rho\mid \bm{y}\right)  = p\left( \bm{y},\bm{a},\bm{w},\bm{\lambda},\tau,\rho\right) /p\left( \bm{y}\right) .
\end{equation}
However, the margin distribution	
\begin{equation}
\begin{aligned}
p\left( \bm{y}\right)
& =\int p\left( \bm{y}\mid \bm{a},\bm{w},\bm{\lambda},\tau,\rho\right) p(\bm{a},\bm{w},\bm{\lambda},\tau,\rho) \ {\rm d}\bm{a}{\rm d}\bm{w}{\rm d}\bm{\lambda}{\rm d}\tau{\rm d}\rho\\
&=\int p\left( \bm{y}\mid \bm{a},\bm{w},\tau\right) p(\bm{a}|\rho)p(\bm{w}|\bm{\lambda})p(\bm{\lambda})p(\tau)p(\rho) \ {\rm d}\bm{a}{\rm d}\bm{w}{\rm d}\bm{\lambda}{\rm d}\tau{\rm d}\rho
\end{aligned}
\end{equation}
is computationally infeasible, because our model is complicated. We have to seek an alternative so as to dispense with the computation of $p\left( \bm{y}\right)$. In the literature of Bayesian methods \cite{BayesLasso,VS,Bishop2006}, there are mainly two types of algorithms to deal with the inference of complex posterior distributions. One is {\it MCMC sampling} which approximates the posterior distribution $p\left( \bm{a},\bm{w},\bm{\lambda},\tau,\rho\mid \bm{y}\right)$ via iteratively drawing samples from the full conditional distributions of each variable. Although MCMC sampling behaves very well in many cases, it is time-consuming when the number of unknown variables is large (i.e., large $N$). The other one is {\it approximation-based techniques} \cite{VB,VB2,VB3} such as variational Bayes or expectation propagation (EP) which works by directly utilizing another easily estimated distribution to approximate the desired posterior distribution. The prominent advantage of these methods is their good performance at the much lower computational cost.

Here, we employ a variational Bayesian method to infer this model. The basic idea is to use a {\it variational distribution} $q(\bm{a},\bm{w},\bm{\lambda},\tau,\rho)$ to approximate the posterior one. The Kullback-Leibler (KL) divergence is utilized to measure the difference between two distributions \cite{VB}. Hence, the original problem is converted into the following optimization issue, namely,
\begin{equation}\label{eq8}
q^*=\min_{q(\bm{a},\bm{w},\bm{\lambda},\tau,\rho)} KL\left( q(\bm{a},\bm{w},\bm{\lambda},\tau,\rho) || p(\bm{a},\bm{w},\bm{\lambda},\tau,\rho\mid \bm{y})\right) .
\end{equation}
Specifically, we hypothesize that the variational distributions for each item are independent, that is,
\begin{equation}\label{eq:variational_distribution}
q(\bm{a},\bm{w},\bm{\lambda},\tau,\rho)=q(\bm{w})q(\rho)q(\tau)\prod_{j=1}^{N}q(\lambda_{j})q(a_j).
\end{equation}
Then, the optimal variational posterior distribution can be obtained by the Theorem \ref{thm1}.

\begin{theorem}\label{thm1}
	The optimal variational posterior distributions of our model are
	\begin{equation}\label{eq9}
	\begin{aligned}
	q(\bm{w}) &= \mathcal{N}(\bm{w}\mid \bm{\mu}, \bm{\Sigma}),\\
	q(\tau) &= {\rm Gamma} (\tau \mid c,d),\\
	q(\rho) &= {\rm Beta} (\rho \mid e,f),\\
	q(a_{j}) &= {\rm Bernoulli} (a_{j}\mid \theta_{j}), j=1,\cdots,N,\\
	q(\lambda_{j}) &= {\rm Gamma}(\lambda_{j}\mid g_j,h_j), j=1,\cdots,N,
	\end{aligned}
	\end{equation}
	where
	\begin{equation}\label{eq:update}
	\begin{aligned}
	\bm{\Omega} &= \bm{\theta}\bm{\theta}^{\rm T}+\mathbb{D}(\bm{\theta})\odot(\bm{I}_N-\mathbb{D}(\bm{\theta})), \\
	\bm{\Sigma}&=\left[\frac{c}{d}(\bm{X}^{\rm T}\bm{X})\odot\bm{\Omega} +\mathbb{D}\left( \frac{\bm{g}}{\bm{h}}\right)  \right] ^{-1}, \quad \bm{\mu} = \frac{c}{d}\bm{\Sigma}\mathbb{D}(\bm{\theta})\bm{X}^{\rm T}\bm{y},\\
	g_j &= g_0+\frac{1}{2},\\
	h_j &= h_0+\frac{1}{2}(\Sigma_{jj}+\mu_j^2),\\
	c &= c_0 +\frac{M}{2}, \\
	d &= d_0 + \frac{1}{2}\left\{ \|\bm{y}\|^2 -2\bm{y}^{\rm T}\bm{X} \mathbb{D}(\bm{\theta})\bm{\mu}+{\rm trace}\left[\left( (\bm{X}^{\rm T}\bm{X}) \odot \bm{\Omega}\right)(\bm{\Sigma}+\bm{\mu}^{\rm T}\bm{\mu})\right]\right\} ,\\
	\theta_{j} &= \frac{1}{\exp(-u_{j})+1}, \\
	u_{j}&= \psi(e)-\psi(f)+\frac{c}{2d}\left\{ X_j^{\rm T}X_j [\mu_j^2\mathbb{D}(\bm{\theta})-0.5(\Sigma_{jj}+\mu_j^2) ] + \mu_jX_j^{\rm T}(\bm{y}-\bm{X}\mathbb{D}(\bm{\theta})\bm{\mu})\right\}, \\
	e &= e_0 + \sum_{j=1}^{N} \theta_{j},f = f_0 + \sum_{j=1}^{N} (1-\theta_{j}).\\
	\end{aligned}
	\end{equation}
	
\end{theorem}
\begin{pf}
	Please see \textbf{Appendix A} for details.
\end{pf}

\subsection{Algorithm and implementation details}
In this part, we describe how to apply our model to network reconstruction tasks. Algorithm 1 lists the main steps of the inference process of our model. In the non-informative fashion \cite{Bishop2006}, the hyper-parameters are initialized as shown in line 1. Algorithm 2 summarizes the workflow of Bayesian complex network reconstruction, that is, repeatedly computing the response vector and the design matrix and then carrying out Algorithm 1 until all nodes' structures are recovered. At last, the output of Algorithm 2, $\hat{\bm{W}}$, is the final reconstructed network. In what follows, we abbreviate our method (Variational Bayesian Reconstruction) as VBR.

\begin{algorithm}[H]
	\caption{Inference of model (\ref{eq:joint_distribution}): $(\bm{\theta},\bm{\mu})=\mathsf{vbr}(\bm{X}, \bm{y})$}
	\label{alg:1}
	\begin{algorithmic}[1]
		\REQUIRE $\bm{X}, \bm{y}$
		\STATE Initialize $g_0=c_0 =10^{-2}, d_0 = h_0=10^{-4}$, $e_0=f_0=1$, $t = 1, \bm{\theta}^{(0)}=\bm{1}, c=c_0+M/2, g_j = g_0+1/2, (j=1,\cdots,p)$.
		\WHILE {the convergence criterion does not satisfy}
		\STATE Update parameters according to (\ref{eq:update});
		\STATE Let $ t = t + 1 $;
		\ENDWHILE
	\end{algorithmic}
\end{algorithm}

\begin{algorithm}[H]
	\caption{Bayesian complex network reconstruction: $\hat{\bm{W}}=\mathsf{BayesRecon}(V, I)$}
	\label{alg:2}
	\begin{algorithmic}[1]
		\REQUIRE $V, I$
		\FOR {$i=1,2,\cdots,N$}
		\STATE Compute the response vector $\bm{y}^{(i)}=(y_{t_1}^{(i)},\cdots,y_{t_M}^{(i)})^{\rm T}$ and the design matrix $\bm{X}^{(i)}=(x_{j,t_m}^{(i)})_{M\times (N-1)}$ with $x_{j,t_m}^{(i)}=V_i(t_m)-V_j(t_m)$, where $m=1,2,\cdots,M$ and $j=1,2,\cdots,i-1,i+1,\cdots,N$.
		\STATE Apply Algorithm \ref{alg:1} to $(\bm{X}^{(i)}, \bm{y}^{(i)})$ and let $(\bm{\theta}_i,\bm{\mu}_i)=\mathsf{vbr}(\bm{X}^{(i)}, \bm{y}^{(i)})$.
		\ENDFOR
		\STATE Let $\hat{w}_{ij}=\mu_{ij}$ if $\theta_{ij}>0.5$, and 0 otherwise.
	\end{algorithmic}
\end{algorithm}

\section{Experiments} \label{sec3}
In this section, we will carry out experiments to study the behavior of our model. The source code of VBR is available at \url{https://xsxjtu.github.io/Projects/VBR/index.html}. Lasso with 5-fold CV to select its tuning parameter is used as the benchmark algorithm and it is implemented by the built-in function $\texttt{lasso}$ in Statistics and Machine Learning Toolbox. All the experiments are carried out on a computer with Intel Core CPU 3.60 GHz, 8.00 GB RAM and Windows 10 (64-bit) system.

To evaluate the reconstruction accuracy of a method, two metrics, ${\rm TPR}$ (true positive rate) and ${\rm TNR}$ (true negative rate), are employed. 
They are defined as
$$
{\rm TPR} = \frac{\rm TP}{\rm P}=\frac{ \sum_{i=1}^{N}\sum_{j=1}^{N}a_{ij}\hat{a}_{ij} }{ \sum_{i=1}^{N}\sum_{j=1}^{N}a_{ij}},
$$
and
$$
{\rm TNR} = \frac{\rm TN}{\rm N}=\frac{ \sum_{i=1}^{N}\sum_{j=1}^{N}(1-a_{ij})(1-\hat{a}_{ij}) }{ \sum_{i=1}^{N}\sum_{j=1}^{N}(1-a_{ij})},
$$
respectively.
As a matter of fact, TPR is the proportion that existed edges are correctly identified while TNR is the proportion that non-existed edges are correctly excluded. Then, error of connection strength is evaluated by $${\rm Error}=\frac{\sqrt{\sum_{i=1}^{N}\sum_{j=1}^{N}(\hat{w}_{ij}-w_{ij})^2}}{\sqrt{\sum_{i=1}^{N}\sum_{j=1}^{N}w_{ij}^2}}.$$ In the meanwhile, we also utilize the computational time (in seconds) to compare the efficiency of each algorithm.

In the following subsections, we will carry out five groups of experiments. In the first experiment, we study the performance of VBR and lasso on Barabasi-Albert (BA) \cite{barabasi1999emergence} and Watts-Strogatz (WS) \cite{watts1998collective} models. BA and WS are typical scale-free and small-world network models, respectively. Generally speaking, small-world networks have large cluster coefficient and small path length; the degree of scale-free networks follows the power-law distribution. Since the exponential-law coefficient of BA networks is a fixed value, in the second experiment, the random scale-free network model with adjustable exponential-law coefficient is employed. As for the third experiment, it aims at investigating the relationship between execution time and network scale. In the forth experiment, we apply VBR and lasso to four real-world networks rather than simulated ones. In the last experiment, we attempt to reconstruct a stock network based the opening price data. 

\subsection{Experiment 1: Performance on BA and WS networks}
In this first case, we mainly compare the performance of VBR and lasso with regard to reconstruction accuracy and running speed on BA and WS networks with ECT and communication dynamics. The aim is to study their behaviors in the cases with different scales of noise. The detailed experimental settings can be found in \textbf{Appendix C}. 

The results are shown in Figs. \ref{fig:ba50} and \ref{fig:ws50}. From left to right, the mean values of TPR, TNR, Error and the time consumed by each algorithm are plotted as a function of $\sigma$, respectively. In each panel, the bar indicates the mean plus or minus one standard deviation. As a matter of fact, it is not surprising that the performance of both VBR and lasso weakens as $\sigma$ increases. Additionally, the following conclusions can be drawn: (1) For lasso, the fluctuation of TNR and Error curves is much greater. So, compared with VBR, lasso is more sensitive to the scale of noise $\sigma$. The reason is that, VBR uses the full Bayesian inference and the variance of noise is directly modeled by the latent variable $\tau$, while lasso does not consider this factor. Besides, this phenomenon is partially caused by lasso's biased estimation property. (2) It is found that in most cases the standard deviation of lasso is higher than that of VBR. It means that lasso may be instable. (3) As for computational time, VBR takes a great advantage over lasso and its speed is very robust to $\sigma$. However, the computational time of lasso dramatically increases as $\sigma$ becomes larger. In general, VBR consumes around 0.05-0.2s and lasso takes 2-21s. 

As a result, in the context of BA and WS networks, VBR outperforms lasso in both reconstruction accuracy and running speed.
\begin{figure}
	\centering
	\includegraphics[width=\linewidth]{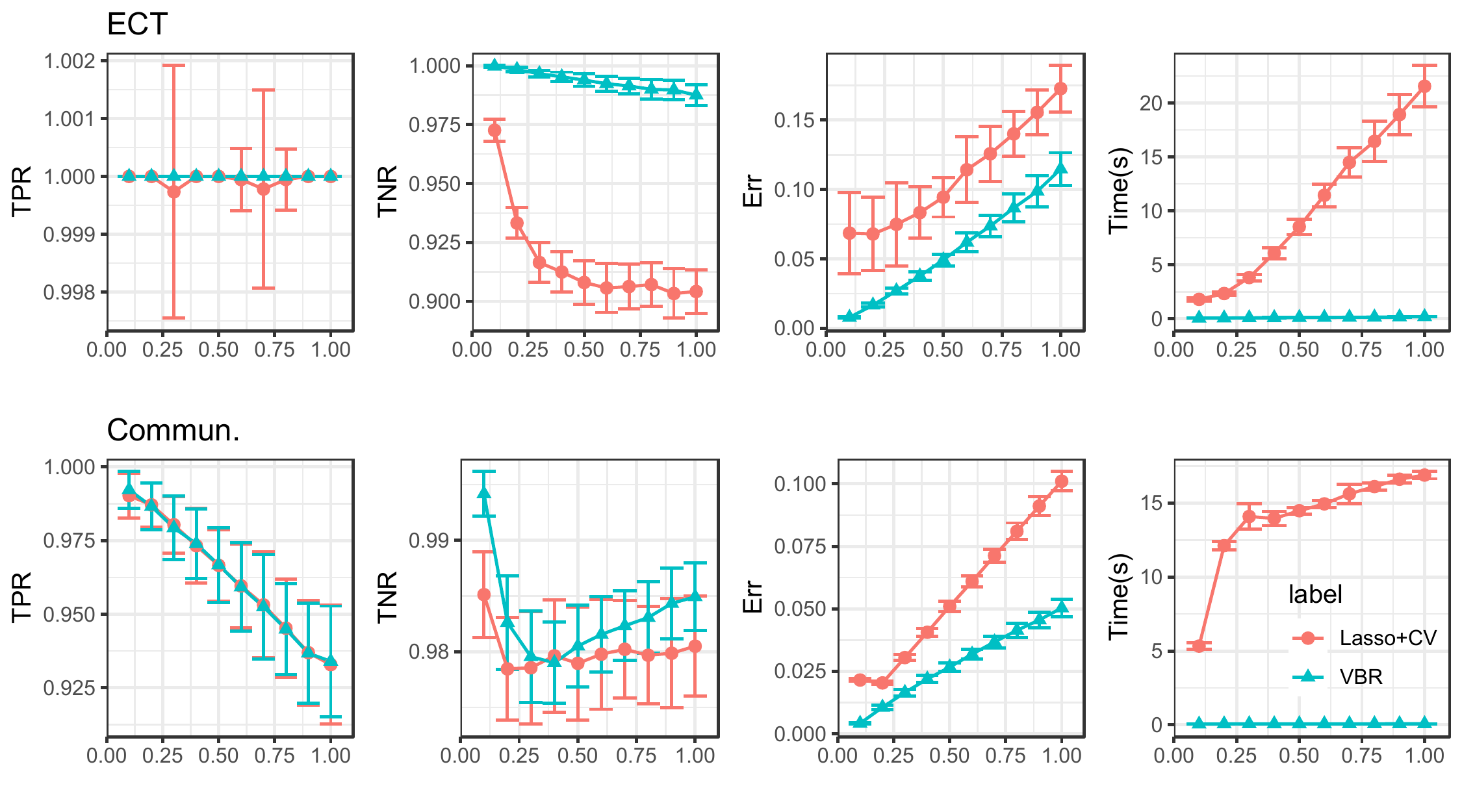}
	\caption{The performance on BA networks (Experiment 1). The experiments were conducted 100 times. The marker and bar denote the mean and standard deviation, respectively. The horizontal axis denotes the scale of noise $\sigma$.}
	\label{fig:ba50}
\end{figure}
\begin{figure}
	\centering
	\includegraphics[width=\linewidth]{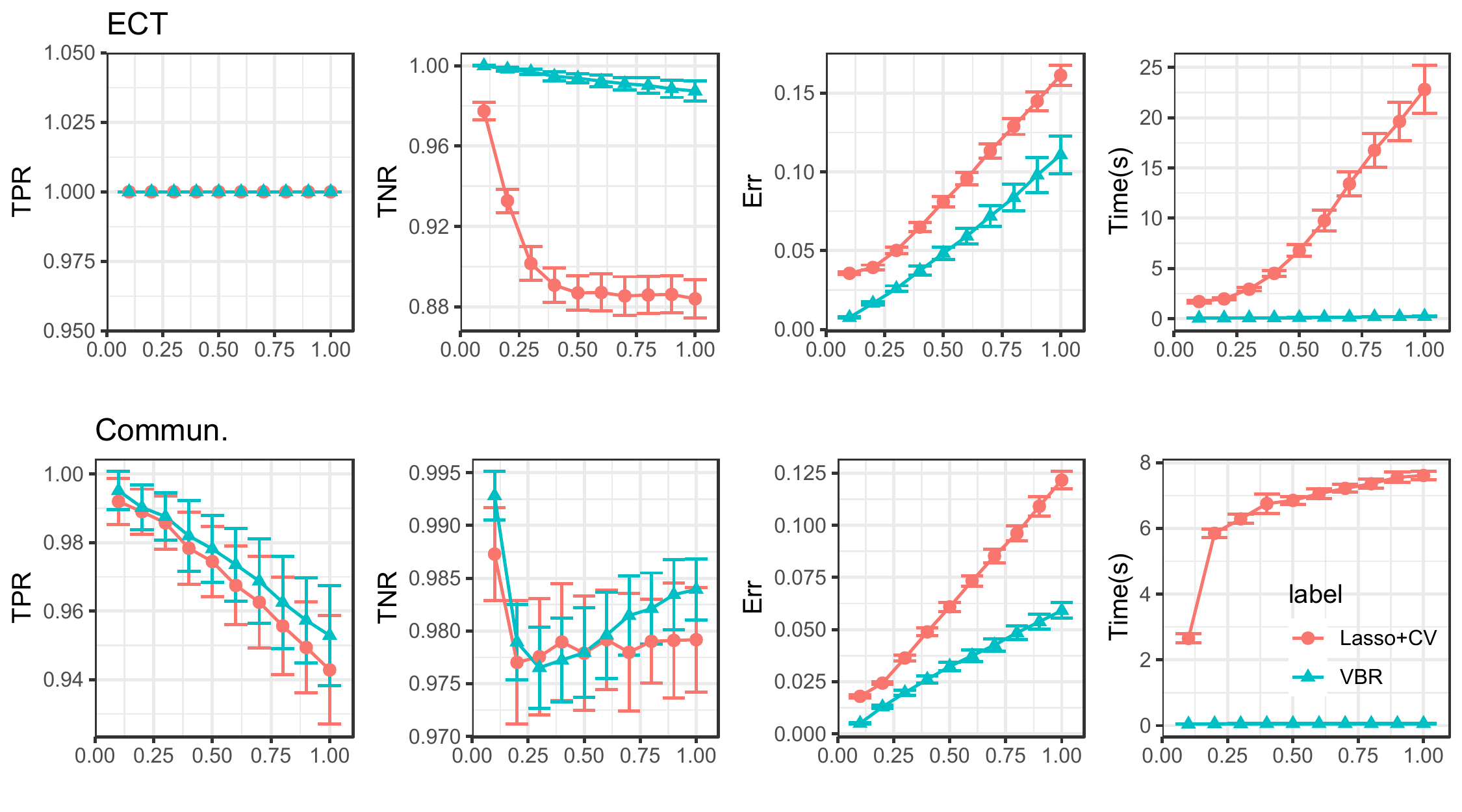}
	\caption{The performance on WS networks (Experiment 1). The experiments were conducted 100 times. The marker and bar denote the mean and standard deviation, respectively. The horizontal axis denotes the scale of noise $\sigma$.}
	\label{fig:ws50}
\end{figure}
\begin{figure}
	\centering
	\includegraphics[width=\linewidth]{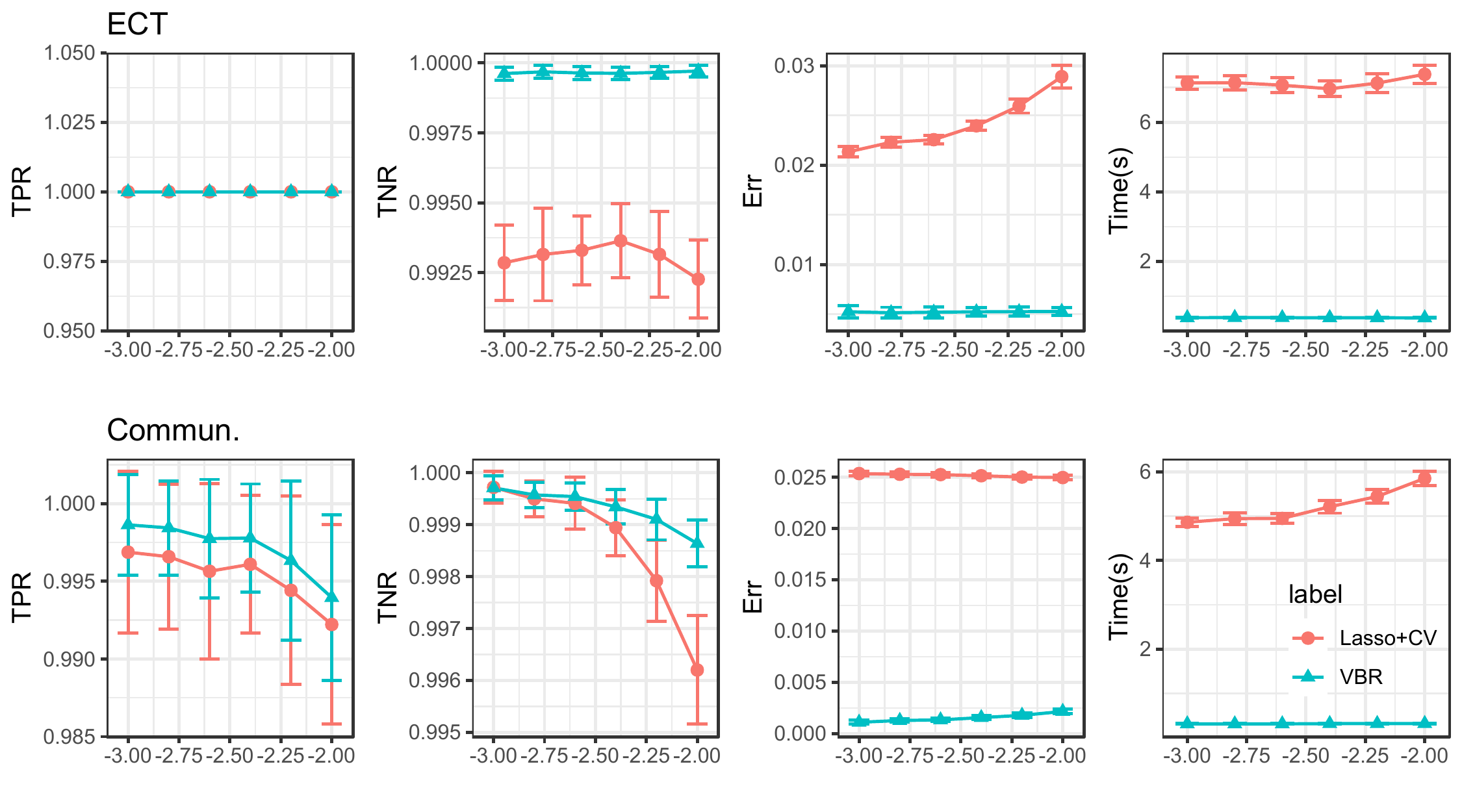}
	\caption{The performance on random SF networks (Experiment 2). The experiments were conducted 100 times. The marker and bar denote the mean and standard deviation, respectively. The horizontal axis denotes exponential-law coefficient $\gamma$.}
	\label{fig:exppow100}
\end{figure}
\subsection{Experiment 2: Performance on random scale-free networks}
The previous investigations have shown that more heterogeneous networks are harder to be reconstructed \cite{Recon_EM,Recon_EM2}. Note that the degree of a scale-free network follows the scale-free distribution, i.e.,
$
p({\rm degree}=k) \propto k^{\gamma},
$
where exponential-law coefficient $\gamma(<0)$ typically ranges from $ -3 $ to $ -2 $. In general, the smaller $|\gamma|$ is, the more heterogeneous a network is. Specifically, the exponential-law coefficient of a BA network is $ -3 $. In this part, we aim to study the methods' behaviors on scale-free networks with adjustable $\gamma$. The detailed experimental settings can be found in \textbf{Appendix C}. 

Fig. \ref{fig:exppow100} reports the results. Firstly, it is shown that VBR outperforms lasso with different exponential-law coefficients in terms of all metrics. Secondly, both VBR and lasso perform worse as $|\gamma|$ becomes smaller. This phenomenon coincides with the former investigations \cite{Recon_EM,Recon_EM2}. However, Fig. \ref{fig:exppow100} shows that the pace of change for lasso is greater than that for VBR. For example, with communication dynamics, the TNR curve of lasso decreases significantly faster when $|\gamma|$ is going smaller. It indicates lasso tends to mistakenly identify null edges as active ones if the scale-free network is more heterogeneous. In conclusion, VBR is more robust to $|\gamma|$ than lasso.

\subsection{Experiment 3: Running speed versus network scale}
Here, we conduct experiments on relatively large-scale complex networks. Because lasso with CV is extremely slow in this case, we only implement VBR. And, the aim is to study how the running speed of VBR varies with the growing of network scale. 

\begin{figure}
	\centering
	\includegraphics[width=0.5\linewidth]{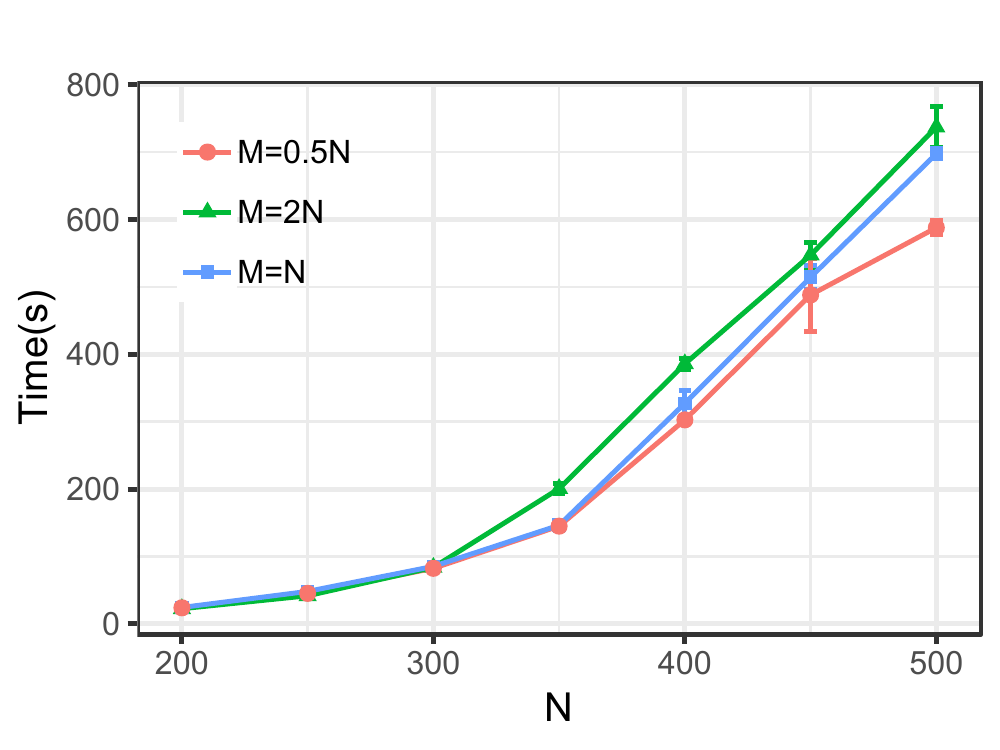}
	\caption{The running speed of VBR with different network scale (Experiment 3). The experiments were conducted 100 times. The marker and bar denote the mean and standard deviation, respectively. BA networks and ECT dynamics are employed.}
	\label{fig:case2}
\end{figure}

Fig. \ref{fig:case2} depicts the execution time of VBR versus $N$, an index indicating the network scale. It can be found in Fig. \ref{fig:case2} that the running speed of VBR grows faster than network scale. The main crux lies in the update of $\bm{w}$, which requires the computation of the inverse of an $N$-dimensional matrix. In general, its computation complexity is $O(N^3)$. Note that it is able to avoid the intensive computation if we assume that $w_j(j=1,2,\cdots,N)$ are independent in Eq. (\ref{eq:variational_distribution}). However, we find that, in our experiments, this strategy is very likely to be tapped into the local minimum, and the values of TPR and TNR are small. Unfortunately, there is still no efficient approach to reduce the computation complexity and it is extremely interesting to study this issue in the future. At last, it is worth pointing out that the computation complexity of lasso is also $O(N^3)$ if it is solved by alternating direction multipliers method.

\subsection{Experiment 4: Performance on real-world networks}
Here, we will conduct experiments on four real-world networks. Since there is no available network with the observed design matrix and response vector, we still simulate them in this subsection. The aim of this experiment is to study the performance of VBR and lasso on real networks rather than the simulated ones (that is, BA, WS and SF networks). HB494BUS and HB1138BUS are available at \url{https://www.cise.ufl.edu/research/sparse/matrices/HB/index.html}; Jazz and Karate are available at \url{http://konect.uni-koblenz.de/networks/}. In this experiment, we did not include noise item and set $\sigma=0$. 

The results reported in Table \ref{tab:2} show that VBR always achieves higher TNR and lower Error than lasso. The values of TPR also manifest the superiority of VBR over lasso. Meanwhile, VBR is very efficient with regard to the execution time. For example, on HB1138BUS, lasso takes about one hour to accomplish the reconstruction task. As for our method VBR, it consumes only less than half an hour to recover the whole network. In summary, the experiments conducted with real-world topology still reveals the superiority of VBR over lasso in terms of all the metrics.

\begin{table}
	\centering
	\caption{Results on empirical networks (Experiment 4). ECT is simulated on the first two networks and communication is simulated on the rest ones.}
	\begin{tabular}{ccccccc}
		\toprule
		Networks&$N$&Method&TPR&TNR&Error&Time\\
		\midrule
		HB494BUS & 494   & Lasso & 0.972  & 1.000  & 0.028  & 238.549  \\
		&       & VBR   & 0.987  & 1.000  & 0.002  & 73.710  \\
		HB1138BUS & 1138  & Lasso & 0.948  & 1.000  & 0.027  & 3381.924  \\
		&       & VBR   & 0.977  & 1.000  & 0.001  & 1553.495  \\
		Jazz  & 198   & Lasso & 0.981  & 0.998  & 0.031  & 49.361  \\
		&       & VBR   & 0.969  & 1.000  & 0.003  & 0.913  \\
		Karate & 34    & Lasso & 0.983  & 0.999  & 0.031  & 0.665  \\
		&       & VBR   & 0.997  & 1.000  & 0.000  & 0.015  \\
		\bottomrule
	\end{tabular}%
	\label{tab:2}%
\end{table}%

\subsection{An empirical study on stock network}
In the last experiment, we apply the reconstruction algorithm to a stock market. Tse et al. state that ``the fluctuations of stock prices are not independent, but are highly inter-coupled with strong correlations with the business sectors and industries to which the stocks belong'' \cite{TSE2010659}. The financial economists have made attempts to analyze the stock market by means of methodologies of complex network. We assume that the price of a stock is the combination of the prices of other stocks and the correlation strengths between stocks do not change in a short period, that is, 
\begin{equation}
s_{it} = \sum_{k\ne i} a_{ik} w_{ik} s_{kt}+\epsilon_{t},
\end{equation}
where $s_{it}$ denotes the price of the $i^{\rm th}$ stock at time $t$, binary variable $a_{ik}$ denotes whether the $k^{\rm th}$ stock affects the $i^{\rm th}$ stock, and $w_{ik}$ is the strength of influence. And $\epsilon_{t}$ is the factor that is independent on $s_{kt}(k=1,2,\cdots,i-1,i+1,\cdots,N)$.

To reconstruct the stock network, we collect the data of stocks in Shanghai Stock Exchange and Shenzhen Stock Exchange from 2018-11-08 to 2019-09-27. It contains the opening prices of 50 stocks during 212 trading days. Each stock is assigned with a label $l_i$, that is, the industry to which it belongs. In this dataset, there are five industries, including agriculture, finance \& insurance, information technology, real estate and transportation. Since there is no ground truth for the network topology, it is hard to evaluate the performances of VBR and lasso. Based on the prior knowledge that the stocks in the same industry tend to correlate with each other \cite{TSE2010659}, we define the cohesion index (CI) for each node as follows,
\begin{equation}
{\rm CI}_i = \frac{\sum_{k\ne i} (a_{ik}+a_{ki}) I(l_i=l_k)}{\sum_{k\ne i} (a_{ik}+a_{ki}) I(l_i\ne l_k)}.
\end{equation}
The numerator (or denominator) of CI counts the number of linkages between node $i$ and the one that belongs to the same (or different) industry. Therefore, greater CI indicates that the node tends to connect with ones from the same industry, and, in other words, the reconstructed network is better. Besides CI, 
we apply a state-of-the-art community detection algorithm, nonnegative matrix factorization, to the reconstructed network, and then use normalized mutual information (NMI) to assess whether the reconstructed network satisfies the prior knowledge. The NMI is defined by 
\begin{equation}
\mathrm{NMI} = \frac{2I(\hat{\bm{l}}; \bm{l})}{H(\hat{\bm{l}})+H(\bm{l})},
\end{equation}	
where $I(\hat{\bm{l}}; \bm{l})=\sum_{\hat{l}\in\mathcal{L}}\sum_{l\in\mathcal{L}} p(\hat{l},l)\log [ p(\hat{l},l)/p(\hat{l})p(l)]$ denotes the mutual information between the true industry label $\bm{l}$ and the cluster label $\hat{\bm{l}}$,  $\mathcal{L}$ is the label space, and $H(\bm{l})=-\sum_{l\in\mathcal{L}}p(l)\log p(l)$ denotes the entropy.
Note that the initialization of nonnegative matrix factorization may affect the result of community detection. To exclude this randomness, we apply nonnegative matrix factorization 100 times and compute the average NMI. The algorithm with larger NMI is better.

\begin{figure}
	\centering
	\includegraphics[width=1\linewidth]{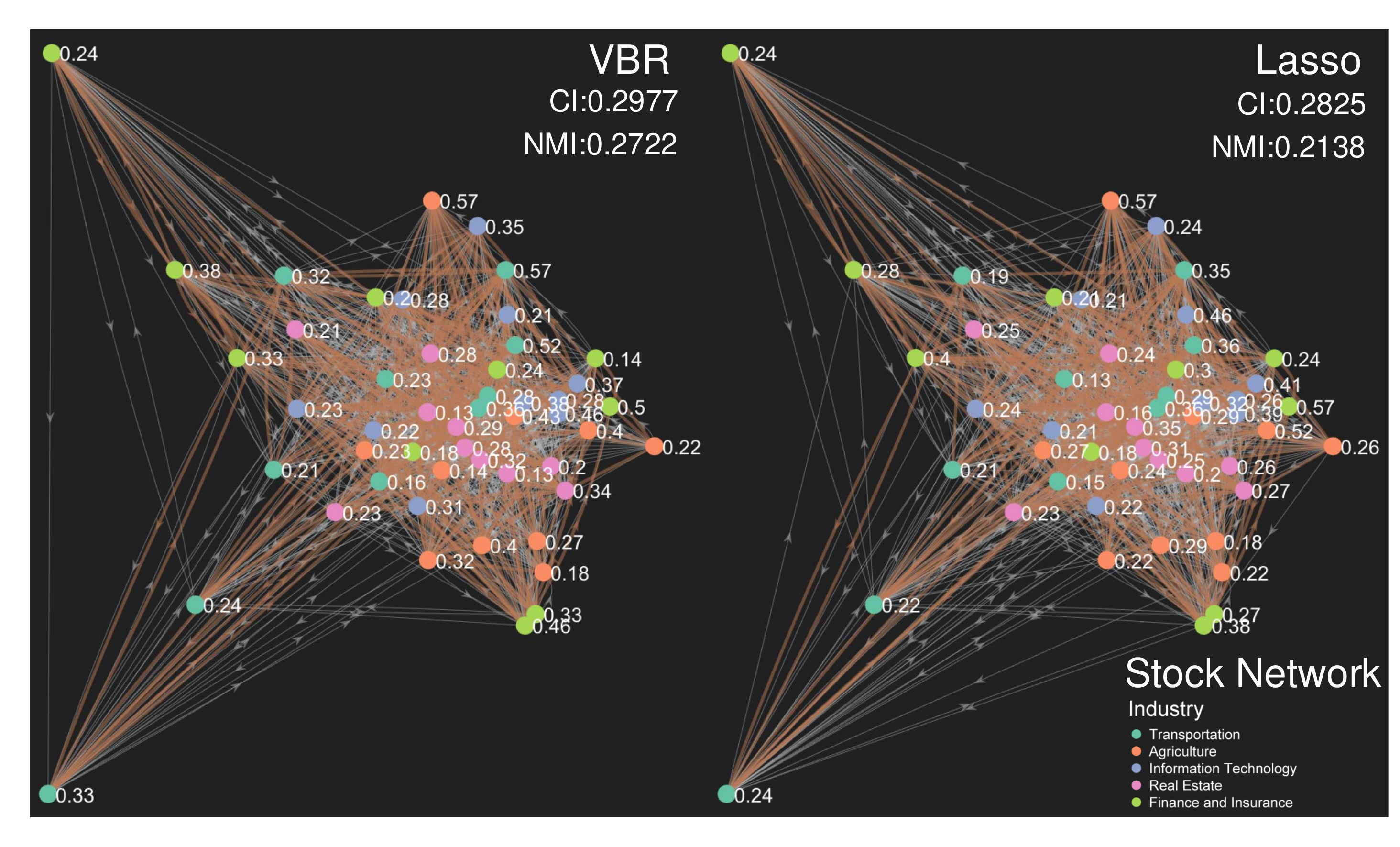}
	\caption{The reconstructed stock network (Experiment 5).}
	\label{fig:stock}
\end{figure}

Fig. \ref{fig:stock} displays the reconstructed stock networks. Each node is colored and labeled by its industry and CI value, respectively. The edge is colored by orange if it connects two nodes from the same industry and gray otherwise. The mean CI of VBR and lasso are 0.2977 and 0.2825, respectively. The NMI of VBR and lasso are 0.2722 and 0.2138, respectively. Both metrics demonstrate that the network recovered by VBR is better in the sense of whether the reconstructed network satisfies the prior knowledge.

\section{Conclusion and future works} \label{sec4}

In this paper, we propose a general framework based on Bayesian statistics to reconstruct weighted complex networks. By reformulating the task as a series of regression problems, prior distributions are assigned to the unknown parameters. To efficiently infer from their posterior distributions, a variational Bayesian method is employed. Compared with lasso, the novel method does not need the fine tuning of its associated parameters. The experiments conducted with both synthetic and real networks show that, in the presence of noise, our method VBR outperforms lasso with regard to both reconstruction accuracy and running speed. 

Finally, we would like to point out the future work. (a) As shown in Eq. (\ref{eq:update}), the update of $\bm{w}$ requires the computation of the inverse of an $N\times N$ matrix. The cost will be unacceptable if $N$ (that is, the number of nodes) is large. At the same time, our experiments show that the lasso based methods also suffer from the similar drawback. With the development of big data era, designing a fast network reconstruction algorithm is one of the key future works. (b) To the best of our knowledge, all papers evaluate the effectiveness of algorithm by simulated data (the design matrix and the response vector). To make the comparisons fairer, it is necessary to create and release a dataset with the observed design matrix, response vector, and the ground truth network topology as a benchmark. (c) Last but not least, both VBR and Lasso cannot handle networks with loops. It is worth paying attention to this problem.

\appendix
\section{Proof of theorem 1}
In this part, we provide more details about the variational inference (VI), which includes a brief introduction of VI and how the variational distributions for the items shown in Theorem 1 are derived.
\subsection{Variational inference}
At first, we show how to use VI to infer the optimal solution for a general model. Let $\bm{z}$ denote all the variables to be inferred, where $z_j$ is the $j^{\rm th}$ variable; and $\bm{y}$ represents the observed data. According to the main principle of VI, the optimal variational distribution is given by
\begin{equation}
q^*(\bm{z}) = \min KL(q(\bm{z})\|p(\bm{z}\mid \bm{y})).
\end{equation}
According to the definition of KL divergence, we have
\begin{equation}\label{eqa1}
\begin{aligned}
KL(q(\bm{z})||p(\bm{z}\mid \bm{y}))
=& \log p(\bm{y}) +
\mathbb{E}_{q(\bm{z})}[\log q(\bm{z})-\log p(\bm{z},\bm{y})].
\end{aligned}
\end{equation}
We define the evidence lower bound (ELBO) as ${\rm ELBO}=\mathbb{E}_{q(\bm{z})}[\log p(\bm{z}, \bm{y})-\log q(\bm{z})]$, where $\mathbb{E}_{q(\bm{z})}$ represents the expectation operator with regard to (w.r.t.) variational distribution $q(\bm{z})$. Note that Eq. (\ref{eqa1}) can be reformulated as
\begin{equation}
\log p(\bm{y}) = KL(q(\bm{z})||p(\bm{z}\mid \bm{y})) +{\rm ELBO}.
\end{equation}
Because $\log p(\bm{y})$ is a constant which does not depend on the variational distribution $q(\bm{z})$, minimizing KL divergence is thus equivalent to maximizing ELBO. Because we assume variational distributions are independent, ELBO can be rewritten as
\begin{equation}\label{eqa4}
{\rm ELBO} 
= -\int \prod_{j}q(z_j) \log \frac{\prod_{j}q(z_j)}{p(\bm{z},\bm{y})} {\rm d}\bm{z} 
- \int q(z_j) \{\log q(z_j) - \mathbb{E}_{-q(z_j)} \log p(\bm{z},\bm{y}) \} {\rm d}z_j + {\rm const.},
\end{equation}
where $\mathbb{E}_{-q(z_j)}$ represents the expectation operator w.r.t. all variational distributions but $q(z_j)$. The ${\rm const.}$ refers to all items that do not depend on $z_j$. Hence, the optimal variational distribution for $z_j$ should satisfy
\begin{equation}\label{eqa5}
\log q^*(z_j) = \mathbb{E}_{-q(z_j)} \left[ \log p(\bm{z},\bm{y})\right].
\end{equation}
Because of this property, the variational distribution $q(\bm{z})$ can be efficiently attained by coordinately updating $q(z_j)$. Remark that $p(\bm{z},\bm{y})$ is the joint distribution. For our model, it refers to Eq. (\ref{eq:joint_distribution}).
In the next, we show how to acquire the optimal solution to our model.

\subsection{Inference of $\bm{w}$}
According to Eq. (\ref{eqa5}), we have
\begin{equation} \label{eq:vd_w}
\begin{aligned}
\log q^*(\bm{w})
=& \mathbb{E}_{-q(\bm{w})} \left\lbrace \log p\left( \bm{y},\bm{a},\bm{w},\bm{\lambda},\tau,\rho\right)  \right\rbrace \\
=& \mathbb{E}_{-q(\bm{w})} \left\lbrace \log p\left( \bm{y}\mid\bm{a},\bm{w},\tau\right)+\log p\left(\bm{w\mid\bm{\lambda}} \right)   \right\rbrace + {\rm const.} \\
=& \mathbb{E}_{-q(\bm{w})} \left\lbrace -\frac{\tau}{2}\|\bm{y}-\bm{X}\mathbb{D}(\bm{a})\bm{w}\|^2 - \frac{1}{2}\bm{w}^{\rm T}\mathbb{D}(\bm{\lambda})\bm{w} \right\rbrace+ {\rm const.} \\
=& \mathbb{E}_{-q(\bm{w})} \left\{ -\frac{\tau}{2}\left[ \bm{w}^{\rm T}\mathbb{D}(\bm{a})\bm{X}^{\rm T}\bm{X}\mathbb{D}(\bm{a})\bm{w}-2\bm{y}^{\rm T}\bm{X}\mathbb{D}(\bm{a})\bm{w}\right] - \frac{1}{2}\bm{w}^{\rm T}\mathbb{D}(\bm{\lambda})\bm{w} \right\}+ {\rm const.} \\
=& \mathbb{E}_{-q(\bm{w})} \left\{ -\frac{1}{2}\bm{w}^{\rm T}\left[\tau\mathbb{D}(\bm{a})\bm{X}^{\rm T}\bm{X}\mathbb{D}(\bm{a})+\mathbb{D}(\bm{\lambda}) \right] \bm{w} +\tau \bm{y}^{\rm T} \bm{X}\mathbb{D}(\bm{a})\bm{w} \right\} +{\rm const.}\\
\end{aligned}
\end{equation}
From Eq. (\ref{eq:vd_w}), it can be seen that the $q(\bm{w})$ is still a Gaussian distribution. In what follows, we denote it by $\mathcal{N}(\bm{w}\mid \bm{\mu},\bm{\Sigma})$ with
\begin{equation}
\begin{aligned}
\bm{\Sigma}&= \left( \mathbb{E}_{q}[\tau] \mathbb{E}_q[\mathbb{D}(\bm{a})\bm{X}^{\rm T}\bm{X}\mathbb{D}(\bm{a})]+\mathbb{E}_q[\mathbb{D}(\bm{\lambda})]\right) ^{-1},\\
\bm{\mu}&= \mathbb{E}_{q}[\tau] \bm{\Sigma} \mathbb{E}_q[\mathbb{D}(\bm{a})] \bm{X}^{\rm T}\bm{y}.\\
\end{aligned}
\end{equation}
It should be mentioned that in $\bm{\mu}$ and $\bm{\Sigma}$, the expectations w.r.t. their corresponding variational distribution will be computed in subsection ``\textbf{The expectation computations}''.

\subsection{Inference of $\lambda_{j}$}
According to Eq. (\ref{eqa5}), there is
\begin{equation}
\begin{aligned}
\log q^*(\lambda_{j})
=& \mathbb{E}_{-q(\lambda_j)} \left\lbrace \log p\left( \bm{y},\bm{a},\bm{w},\bm{\lambda},\tau,\rho\right)  \right\rbrace \\
=&\mathbb{E}_{-q(\lambda_j)} \left\lbrace \log p(w_j\mid\lambda_j)+\log p(\lambda_j)\right\rbrace +{\rm const.}\\
=&\mathbb{E}_{-q(\lambda_j)} \left\{ \frac{1}{2}\log\lambda_j -\frac{\lambda_j}{2} w_j^2+(g_0-1)\log\lambda_j- h_0 \lambda_j\right\} +{\rm const.}\\
\end{aligned}
\end{equation}
Therefore, $q(\lambda_j)$ is still a Gamma distribution. In what follows, we denote it by ${\rm Gamma}(g_j,h_j)$, where
\begin{equation}
g_j = g_0+\frac{1}{2}, \quad h_j=h_0+\frac{1}{2} \mathbb{E}_q(w_j^2).
\end{equation}

\subsection{Inference of $\tau$}
As for $\tau$, we have
\begin{equation}
\begin{aligned}
\log q^*(\tau) 
=& \mathbb{E}_{-q(\tau)} \left\lbrace \log p\left( \bm{y},\bm{a},\bm{w},\bm{\lambda},\tau,\rho\right)  \right\rbrace \\
=& \mathbb{E}_{-q(\tau)}\left\lbrace \log p(\bm{y}\mid \bm{a}, \bm{w}, \tau)+\log  p(\tau)\right\rbrace + {\rm const.}\\
=& \frac{M}{2}\log\tau-\frac{\tau}{2}\mathbb{E}_{-q(\tau)}\left[ \|\bm{y}-\bm{X}\mathbb{D}(\bm{a})\bm{w}\|^2\right]  + (c_0-1)\log\tau -(d_0-1)\tau+ {\rm const.}\\
=&\left(c_0+\frac{M}{2}-1\right)\log\tau -(d_0+ \mathbb{E}_{-q(\tau)}\left[ \|\bm{y}-\bm{X}\mathbb{D}(\bm{a})\bm{w}\|^2\right]-1)\tau+{\rm const.}
\end{aligned}
\end{equation}
according to Eq. (\ref{eqa5}). Hence, $q(\tau)$ is still a Gamma distribution. In what follows, we denote it by ${\rm Gamma}(\tau\mid c, d)$, where
\begin{equation}
c = c_0 + \frac{M}{2}, d = d_0 + \frac{1}{2} \mathbb{E}_{q}\left[ \|\bm{y}-\bm{X}\mathbb{D}(\bm{a})\bm{w}\|^2\right].
\end{equation}

\subsection{Inference of $a_{j}$}
To facilitate the derivation, it is worthy mentioning one fact that, if $x\sim{\rm Bernoulli}(\rho)$, then $x^2\sim{\rm Bernoulli}(\rho)$. That is, $x$ and $x^2$ have the identical distribution. According to Eq. (\ref{eqa5}), the optimal variational distribution for $a_{j}$ should satisfy
\begin{equation}
\begin{aligned}
\log q^*(a_{j}) 
=& \mathbb{E}_{-q(\tau)} \left\lbrace \log p\left( \bm{y},\bm{a},\bm{w},\bm{\lambda},\tau,\rho\right)  \right\rbrace \\
=& \mathbb{E}_{-q(a_{j})} \{  -\frac{\tau}{2} \|\bm{y}-\bm{X}\mathbb{D}(\bm{a})\bm{w}\|^2  +a_{j}\log\rho+(1-a_{j})\log(1-\rho) \}+{\rm const.}\\
=&\mathbb{E}_{-q(a_{j})} \left\{-\frac{\tau}{2} \|\bm{y}-\sum_{n\ne j}X_n a_{n}w_n-X_j a_{j}w_j\|^2 +a_{j}\log\frac{\rho}{1-\rho} \right\}+{\rm const.}\\
=&\mathbb{E}_{-q(a_{j})} \left\{-\frac{\tau}{2}\|\bm{r}_j-X_ja_{j}w_j\|^2+a_{j}\log\frac{\rho}{1-\rho} \right\} +{\rm const.}\\
=&\mathbb{E}_{-q(a_{j})}\left\{-\frac{\tau}{2}(-2\bm{r}_j^{\rm T}X_ja_{j}w_j+X_j^{\rm T}X_ja_{j}^2w_j^2 ) +a_{j}\log\frac{\rho}{1-\rho} \right\} +{\rm const.}\\
=&a_{j}\mathbb{E}_{-q(a_{j})}\left\{-\frac{\tau}{2}X_j^{\rm T}(w_j^2X_j-2w_j\bm{r}_j)+\log\frac{\rho}{1-\rho} \right\} +{\rm const.}\\
\end{aligned}
\end{equation}
Here, $X_j$ represents the $j^{\rm th}$ column of matrix $\bm{X}$ and  $\bm{r}_j\equiv\bm{y}-\sum_{n\ne j}X_na_{n}$. Obviously, $q(a_{j})$ is still a Bernoulli distribution. In what follows, we denote it by ${\rm Bernoulli}(\theta_{j})$ with
\begin{equation}
\theta_{j} = \frac{1}{\exp(-u_{j})+1},
u_{j}=\mathbb{E}_{q}\left\lbrace \log\frac{\rho}{1-\rho}-\frac{\tau}{2}X_j^{\rm T}(w_j^2X_j-2w_j\bm{r}_j)\right\rbrace .
\end{equation}

\subsection{Inference of $\rho$}
Similar to Eq. (\ref{eqa4}), the optimal variational distribution for $\rho$ should satisfy
\begin{equation}
\begin{aligned}
\log q(\rho) 
=& \mathbb{E}_{-q(\rho)} \left\lbrace \log p(\rho)+\sum_{j=1}^{N}\log p(a_{j}\mid \rho)\right\rbrace+{\rm const.}\\
=&  \mathbb{E}_{-q(\rho)} \{ (e_0-1)\log\rho+(f_0-1)\log(1-\rho) +\sum_{j=1}^{N}a_{j}\log\rho+(1-a_{j})\log(1-\rho) \}+{\rm const.}\\
=&\left(f_0+\sum_{j=1}^{N}(1-\mathbb{E}_{-q(\rho)}[a_{j}])-1\right) \log(1-\rho)  +\left(e_0+\sum_{j=1}^{N}\mathbb{E}_{-q(\rho)}[a_{j}]-1\right) \log\rho +{\rm const.}
\end{aligned}
\end{equation}
Therefore, $q(\rho)$ is still a Beta distribution. In the current paper, we denote it by ${\rm Beta}(e,f)$, where
\begin{equation}
e = e_0 + \sum_{j=1}^{N}\mathbb{E}_{q}[a_{j}],
f = f_0 + \sum_{j=1}^{N}(1-\mathbb{E}_{q}[a_{j}]).
\end{equation}

\subsection{The expectation computations}\label{sec:app_G}
So far, we have attained the variational distributions for all (hyper)parameters. However, there are some expectations remaining unsolved. In this part, we show how these expectations can be computed. For simplicity, the subscript `$q$' is omitted from $\mathbb{E}_q$ in the following discussions.

\begin{enumerate}
	\item $\mathbb{E} [\tau] = c/d$.
	\item $\mathbb{E}[\bm{a}] = \bm{\theta}=(\theta_{1},\cdots,\theta_{N})^{\rm T}$.
	\item $\mathbb{E}[\mathbb{D}(\bm{\lambda})] = \mathbb{D}(\bm{g}/\bm{h})=\mathbb{D}(g_1/h_1,\cdots,g_N/h_N)$.
	\item Since $\mathbb{E}[a_{j}a_{n}]=\theta_{j}\theta_{n} (j\ne n)$ and $\mathbb{E}[a_{j}^2]=\theta_{j}=\theta_{j}^2+\theta_{j}(1-\theta_{j})$, we can thus obtain that $\mathbb{E}[\bm{a}\bm{a}^{\rm T}] = \bm{\Omega} = \bm{\theta}\bm{\theta}^{\rm T}+\mathbb{D}(\bm{\theta})\odot(\bm{I}_N-\mathbb{D}(\bm{\theta}))$, where $\mathbb{D}(\bm{\theta})={\rm diag}(\bm{\theta})$ and $\bm{I}_N$ is an $N$-dimensional identity matrix.
	\item Note that there is $\bm{w} \sim \mathcal{N}(\bm{w}\mid \bm{\mu},\bm{\Sigma})$. Hence, we have $(\bm{w}-\bm{\mu})(\bm{w}-\bm{\mu})^{\rm T}\sim \mathrm{Wishart}(1,\bm{\Sigma})$, where 1 denotes the degree of freedom and $\bm{\Sigma}$ is the location parameter of the Wishart distribution. Then, there is
	$$
	\mathbb{E}[(\bm{w}-\bm{\mu})(\bm{w}-\bm{\mu})^{\rm T}] = \bm{\Sigma} \Rightarrow \mathbb{E}[\bm{ww}^{\rm T}]=\bm{\Sigma}+\bm{\mu\mu}^{\rm T}.
	$$
	Furthermore, $\mathbb{E}[w_j^2]=\Sigma_{jj}+\mu_j^2$.
	\item According to 4, we have $\mathbb{E}[\mathbb{D}(\bm{a})\bm{X}^{\rm T}\bm{X}\mathbb{D}(\bm{a})] = \mathbb{E}   \left[ \left( \bm{X}^{\rm T}\bm{X}\right) \odot  \left( \bm{a}\bm{a}^{\rm T}\right) \right]   = \left( \bm{X}^{\rm T}\bm{X}\right) \odot \bm{\Omega}$.
	\item According to 5 and 6, we have
	$$
	\begin{aligned}
	\mathbb{E}[\bm{w}^{\rm T}\mathbb{D}(\bm{a})\bm{X}^{\rm T}\bm{X}\mathbb{D}(\bm{a})\bm{w}]=&\mathbb{E}\left\lbrace {\rm trace}\left[ \bm{w}^{\rm T}\mathbb{D}(\bm{a})\bm{X}^{\rm T}\bm{X}\mathbb{D}(\bm{a})\bm{w}\right] \right\rbrace\\
	=& {\rm trace} \left\lbrace \mathbb{E}\left[ \bm{ww}^{\rm T} \mathbb{D}(\bm{a})\bm{X}^{\rm T}\bm{X}\mathbb{D}(\bm{a})\right]  \right\rbrace \\
	=&{\rm trace}\left\lbrace  (\bm{\Sigma}+\bm{\mu\mu}^{\rm T}) \left[ \left( \bm{X}^{\rm T}\bm{X}\right) \odot \bm{\Omega}\right] \right\rbrace .
	\end{aligned} $$
	\item According to 5 and 6, we have
	$$
	\mathbb{E}\left[ \|\bm{y}-\bm{X}\mathbb{D}(\bm{a})\bm{w}\|^2\right] 
	= \|\bm{y}\|^2 -2\bm{y}^{\rm T}\bm{X}  \mathbb{D}(\bm{\theta})\bm{\mu}   +{\rm trace}\left\lbrace  (\bm{\Sigma}+\bm{\mu\mu}^{\rm T}) \left[ \left( \bm{X}^{\rm T}\bm{X}\right) \odot \bm{\Omega}\right] \right\rbrace.
	$$
	\item $\mathbb{E}[\bm{r}_j] = \bm{y}-\sum_{n\ne j}X_n\mu_n\theta_{n}$;
	\item $\mathbb{E}\left[\log\frac{\rho}{1-\rho} \right] =\psi(e)-\psi(e+f)-\psi(f)+\psi(e+f) = \psi(e)-\psi(f)$, where $\psi(x)$ denotes the digamma function defined as the logarithmic derivative of the gamma function.
\end{enumerate}

By plugging these expectations into the variational distributions derived in previous subsections, and the final solutions shown in Theorem 1 can be easily obtained.

\section{Communication dynamics}
In this subsection, we introduce how to apply VBR to communication dynamics described in literature \cite{CS_CN}. The communication dynamics is used to capture communications in populations via phones or Emails. At time $t_m$, individual $i$ contact one of its neighbors $j$ with probability $w_{ji}$ by sending data packets. In this period, the total incoming flux of $i$ is 
\begin{equation}\label{eq:Communication_dynamics}
f_{t_m}^{(i)} = \sum_{j=1}^{N}a_{ji}w_{ji} o_{j, t_m},
\end{equation}
where $o_{j,t_m}$ is the total outgoing flux from $j$ to its neighbors at time $t_m$ and $\sum_{j=1}^{N}a_{ji}w_{ji}=1$. Note the total outgoing flux fluctuates with time. Therefore, we have
\begin{equation} \label{eqx}
\left[ \begin{matrix} f^{(i)}_{t_1} \\ f^{(i)}_{t_2} \\ \vdots \\ f^{(i)}_{t_M}	\end{matrix} \right]  = 	
\left[ \begin{matrix} o_{1,t_1} & o_{2,t_1} & \cdots & o_{N,t_1} \\ o_{1,t_2} & o_{2,t_2} & \cdots & o_{N,t_2} \\ \vdots & \vdots & & \vdots \\ o_{1,t_M} & o_{2,t_M} & \cdots & o_{N,t_M}\end{matrix}\right] 	
\left[ \begin{matrix} a_{1i}w_{1i} \\ a_{2i}w_{2i} \\ \vdots \\ a_{Ni}w_{Ni}	\end{matrix} \right].
\end{equation}
Obviously, VBR can be employed to deal with this case. 

\section{Experimental Settings}
\subsection{Experiment 1}
We mimic the electrical current transportation (ECT) on a network $\bm{W}$ that is described by following equation, 
\begin{equation}
\sum_{j=1}^{N} \frac{a_{ij}}{w_{ij}}(V_i-V_j)=I_i \quad (i=1,2,\cdots,N),
\end{equation}
where the node's voltage is generated by alternating current $V_i=\bar{V}\sin[(\omega+\Delta\omega_i)t]$ with $\bar{V}=1$, $\omega=10^3$ and $\Delta\omega_i$ and the resistance $w_{ij}$ are uniformly sampled from [0, 20] and [2, 3], respectively. 

As for the communication dynamics on a network $\bm{W}$ that is described by Eq. (\ref{eq:Communication_dynamics}), the communication probability $w_{ji}$ is uniformly sampled from [0,1], and then is normalized such that $\sum_{j=1}^{N}a_{ji}w_{ji}=1$. The total outgoing flux $o_{j,t_m}$ is uniformly sampled from [0,20]. 

The variance of the noise item is $\sigma^2$. In the experiments, we consider $\sigma=0.1,0.2,\cdots,1.0$. For each case, the experiment was conducted 100 times by generating different data. The network size $N=50$. For ECT dynamics, we set the number of observations $M=50$. Since network reconstruction of communication dynamics is harder than that of ECT, we set the number of observations $M=200$ for communication dynamics. 

The experiments are conducted on simulated BA and WS networks. The BA networks are generated by the code written by Mathew George \footnote{available at \url{https://ww2.mathworks.cn/matlabcentral/fileexchange/11947-b-a-scale-free-network-generation -and-visualization}}. The WS networks are generated by the official code of Matlab \footnote{available at \url{https://ww2.mathworks.cn/help/matlab/math/build-watts-strogatz-small-world-graph-model.html?lang=en}}.

\subsection{Experiment 2} 
The most settings of Experiment 2 are the same to those of Experiment 1. However, in this part, we set $\sigma=0.1$. The random scale-free network with exponential-law coefficient $\gamma$ is generated by \texttt{mexGraphCreateRandomGraph}, a function of Complex Networks Package \footnote{available at \url{http://www.levmuchnik.net/Content/Networks/ComplexNetworksPackage.html}}. This function is able to generate a graph of given size and with the given node's degree distribution. In the experiments, $\gamma=-2,-2.2,\cdots,-3$. The number of node is $N=100$. 

\section*{Competing interests}
The authors declare that they have no competing interests.

\section*{Acknowledgements}
The authors would like to thank the editor, the associate editor and anonymous reviewers for their useful suggestions which greatly helped to improve the paper. The research of S. Xu is supported by the Fundamental Research Funds for the Central Universities [grant number xzy022019059]. The research of C.X. Zhang is supported by the National Natural Science Foundation of China [grant number 11671317] and the National Key Research and Development Program of China [grant number 2018AAA0102201]. The research of P. Wang is supported by the National Natural Science Foundation of China [grant number 61773153], the Supporting Plan for Scientific and Technological Innovative Talents in Universities of Henan Province [grant number 20HASTIT025], and the Training Plan of Young Key Teachers in Colleges and Universities of Henan Province [grant number 2018GGJS021]. The research of J.S. Zhang is supported by the National Key Research and Development Program of China [grant number 2018YFC0809001], and the National Natural Science Foundation of China [grant number 61976174]. 





\bibliography{ref}

\begin{thebibliography}{41}
\expandafter\ifx\csname natexlab\endcsname\relax\def\natexlab#1{#1}\fi
\providecommand{\url}[1]{\texttt{#1}}
\providecommand{\href}[2]{#2}
\providecommand{\path}[1]{#1}
\providecommand{\DOIprefix}{doi:}
\providecommand{\ArXivprefix}{arXiv:}
\providecommand{\URLprefix}{URL: }
\providecommand{\Pubmedprefix}{pmid:}
\providecommand{\doi}[1]{\href{http://dx.doi.org/#1}{\path{#1}}}
\providecommand{\Pubmed}[1]{\href{pmid:#1}{\path{#1}}}
\providecommand{\bibinfo}[2]{#2}
\ifx\xfnm\relax \def\xfnm[#1]{\unskip,\space#1}\fi
\bibitem[{Arlot and Celisse(2010)}]{CrossValidation}
\bibinfo{author}{Arlot, S.}, \bibinfo{author}{Celisse, A.},
  \bibinfo{year}{2010}.
\newblock \bibinfo{title}{A survey of cross-validation procedures for model
  selection}.
\newblock \bibinfo{journal}{Stat. Surveys} \bibinfo{volume}{4},
  \bibinfo{pages}{40--79}.
\newblock \DOIprefix\doi{10.1214/09-SS054}.
\bibitem[{Barab{\'a}si and Albert(1999)}]{barabasi1999emergence}
\bibinfo{author}{Barab{\'a}si, A.L.}, \bibinfo{author}{Albert, R.},
  \bibinfo{year}{1999}.
\newblock \bibinfo{title}{Emergence of scaling in random networks}.
\newblock \bibinfo{journal}{Science} \bibinfo{volume}{286},
  \bibinfo{pages}{509--512}.
\newblock \DOIprefix\doi{10.1126/science.286.5439.509}.
\bibitem[{Barab\'{a}si and Oltvai(2004)}]{Biology0}
\bibinfo{author}{Barab\'{a}si, A.L.}, \bibinfo{author}{Oltvai, Z.N.},
  \bibinfo{year}{2004}.
\newblock \bibinfo{title}{Network biology: understanding the cell's functional
  organization}.
\newblock \bibinfo{journal}{Nat. Rev. Genet.} \bibinfo{volume}{5},
  \bibinfo{pages}{101--979}.
\newblock \DOIprefix\doi{10.1038/nrg1272}.
\bibitem[{Bishop(2006)}]{Bishop2006}
\bibinfo{author}{Bishop, C.M.}, \bibinfo{year}{2006}.
\newblock \bibinfo{title}{Pattern Recognition and Machine Learning}.
\newblock \bibinfo{publisher}{Springer-Verlag, New York}.
\bibitem[{Blei et~al.(2017)Blei, Kucukelbir and McAuliffe}]{VB}
\bibinfo{author}{Blei, D.M.}, \bibinfo{author}{Kucukelbir, A.},
  \bibinfo{author}{McAuliffe, J.D.}, \bibinfo{year}{2017}.
\newblock \bibinfo{title}{Variational inference: A review for statisticians}.
\newblock \bibinfo{journal}{J. Am. Stat. Assoc.} \bibinfo{volume}{112},
  \bibinfo{pages}{859--877}.
\newblock \DOIprefix\doi{10.1080/01621459.2017.1285773}.
\bibitem[{Brovelli et~al.(2004)Brovelli, Ding, Ledberg, Chen, Nakamura and
  Bressler}]{GrangerRecon}
\bibinfo{author}{Brovelli, A.}, \bibinfo{author}{Ding, M.},
  \bibinfo{author}{Ledberg, A.}, \bibinfo{author}{Chen, Y.},
  \bibinfo{author}{Nakamura, R.}, \bibinfo{author}{Bressler, S.L.},
  \bibinfo{year}{2004}.
\newblock \bibinfo{title}{Beta oscillations in a large-scale sensorimotor
  cortical network: Directional influences revealed by granger causality}.
\newblock \bibinfo{journal}{Proc. Natl. Acad. Sci. U. S. A.}
  \bibinfo{volume}{101}, \bibinfo{pages}{9849--9854}.
\newblock \DOIprefix\doi{10.1073/pnas.0308538101}.
\bibitem[{Donoho(2006)}]{CS}
\bibinfo{author}{Donoho, D.L.}, \bibinfo{year}{2006}.
\newblock \bibinfo{title}{Compressed sensing}.
\newblock \bibinfo{journal}{IEEE Trans. Info. Theory} \bibinfo{volume}{52},
  \bibinfo{pages}{1289--1306}.
\newblock \DOIprefix\doi{10.1109/TIT.2006.871582}.
\bibitem[{Granger(1969)}]{GrangeCausality}
\bibinfo{author}{Granger, C.W.J.}, \bibinfo{year}{1969}.
\newblock \bibinfo{title}{Investigating causal relations by econometric models
  and cross-spectral methods}.
\newblock \bibinfo{journal}{Econometrica} \bibinfo{volume}{37},
  \bibinfo{pages}{424--438}.
\newblock \DOIprefix\doi{10.2307/1912791}.
\bibitem[{Guo et~al.(2008)Guo, Seth, Kendrick, Zhou and
  Feng}]{PartialGrangerCausality}
\bibinfo{author}{Guo, S.}, \bibinfo{author}{Seth, A.K.},
  \bibinfo{author}{Kendrick, K.M.}, \bibinfo{author}{Zhou, C.},
  \bibinfo{author}{Feng, J.}, \bibinfo{year}{2008}.
\newblock \bibinfo{title}{Partial {G}ranger causality--eliminating exogenous
  inputs and latent variables.}
\newblock \bibinfo{journal}{J. Neurosci. Methods} \bibinfo{volume}{172},
  \bibinfo{pages}{79--93}.
\newblock \DOIprefix\doi{10.1016/j.jneumeth.2008.04.011}.
\bibitem[{Han et~al.(2015)Han, Shen, Wang and Di}]{CS_CN}
\bibinfo{author}{Han, X.}, \bibinfo{author}{Shen, Z.}, \bibinfo{author}{Wang,
  W.X.}, \bibinfo{author}{Di, Z.}, \bibinfo{year}{2015}.
\newblock \bibinfo{title}{Robust reconstruction of complex networks from sparse
  data}.
\newblock \bibinfo{journal}{Phys. Rev. Lett.} \bibinfo{volume}{114},
  \bibinfo{pages}{Art. No. 028701}.
\newblock \DOIprefix\doi{10.1103/PhysRevLett.114.028701}.
\bibitem[{Jordan et~al.(1999)Jordan, Ghahramani, Jaakkola and Saul}]{VB2}
\bibinfo{author}{Jordan, M.I.}, \bibinfo{author}{Ghahramani, Z.},
  \bibinfo{author}{Jaakkola, T.S.}, \bibinfo{author}{Saul, L.K.},
  \bibinfo{year}{1999}.
\newblock \bibinfo{title}{An introduction to variational methods for graphical
  models}.
\newblock \bibinfo{journal}{Mach. Learn.} \bibinfo{volume}{37},
  \bibinfo{pages}{183--233}.
\newblock \DOIprefix\doi{10.1023/A:1007665907178}.
\bibitem[{Liu et~al.(2018)Liu, Mei, Wu and L\"{u}}]{CS_weight}
\bibinfo{author}{Liu, J.}, \bibinfo{author}{Mei, G.}, \bibinfo{author}{Wu, X.},
  \bibinfo{author}{L\"{u}, J.}, \bibinfo{year}{2018}.
\newblock \bibinfo{title}{Robust reconstruction of continuously time-varying
  topologies of weighted networks}.
\newblock \bibinfo{journal}{IEEE Trans. Circuits Syst. I-Regul. Pap.}
  \bibinfo{volume}{65}, \bibinfo{pages}{2970--2982}.
\newblock \DOIprefix\doi{10.1109/TCSI.2018.2808233}.
\bibitem[{Ma et~al.(2018)Ma, Chen, Lai and Zhang}]{Recon_EM}
\bibinfo{author}{Ma, C.}, \bibinfo{author}{Chen, H.S.}, \bibinfo{author}{Lai,
  Y.C.}, \bibinfo{author}{Zhang, H.F.}, \bibinfo{year}{2018}.
\newblock \bibinfo{title}{Statistical inference approach to structural
  reconstruction of complex networks from binary time series}.
\newblock \bibinfo{journal}{Phys. Rev. E} \bibinfo{volume}{97},
  \bibinfo{pages}{022301}.
\newblock \DOIprefix\doi{10.1103/PhysRevE.97.022301}.
\bibitem[{Mei et~al.(2018)Mei, Wu, Wang, Hu, Lu and Chen}]{CS_multi}
\bibinfo{author}{Mei, G.}, \bibinfo{author}{Wu, X.}, \bibinfo{author}{Wang,
  Y.}, \bibinfo{author}{Hu, M.}, \bibinfo{author}{Lu, J.},
  \bibinfo{author}{Chen, G.}, \bibinfo{year}{2018}.
\newblock \bibinfo{title}{Compressive-sensing-based structure identification
  for multilayer networks}.
\newblock \bibinfo{journal}{IEEE Trans. Cybernetics} \bibinfo{volume}{48},
  \bibinfo{pages}{754--764}.
\newblock \DOIprefix\doi{10.1109/TCYB.2017.2655511}.
\bibitem[{Nardelli et~al.(2014)Nardelli, Rubido, Wang, Baptista, Pomalaza-Raez,
  Cardieri and Latva-aho}]{PowerGrid2}
\bibinfo{author}{Nardelli, P.H.}, \bibinfo{author}{Rubido, N.},
  \bibinfo{author}{Wang, C.}, \bibinfo{author}{Baptista, M.S.},
  \bibinfo{author}{Pomalaza-Raez, C.}, \bibinfo{author}{Cardieri, P.},
  \bibinfo{author}{Latva-aho, M.}, \bibinfo{year}{2014}.
\newblock \bibinfo{title}{Models for the modern power grid}.
\newblock \bibinfo{journal}{Eur. Phys. J-Spec. Top.} \bibinfo{volume}{223},
  \bibinfo{pages}{2423--2437}.
\newblock \DOIprefix\doi{10.1140/epjst/e2014-02219-6}.
\bibitem[{Onnela et~al.(2007)Onnela, Saram{\"a}ki, Hyv{\"o}nen, Szab{\'o},
  Lazer, Kaski, Kert{\'e}sz and Barab{\'a}si}]{Social1}
\bibinfo{author}{Onnela, J.P.}, \bibinfo{author}{Saram{\"a}ki, J.},
  \bibinfo{author}{Hyv{\"o}nen, J.}, \bibinfo{author}{Szab{\'o}, G.},
  \bibinfo{author}{Lazer, D.}, \bibinfo{author}{Kaski, K.},
  \bibinfo{author}{Kert{\'e}sz, J.}, \bibinfo{author}{Barab{\'a}si, A.L.},
  \bibinfo{year}{2007}.
\newblock \bibinfo{title}{Structure and tie strengths in mobile communication
  networks}.
\newblock \bibinfo{journal}{Proc. Natl. Acad. Sci. U. S. A.}
  \bibinfo{volume}{104}, \bibinfo{pages}{7332--7336}.
\newblock \DOIprefix\doi{10.1073/pnas.0610245104}.
\bibitem[{Pagani and Aiello(2013)}]{PowerGrid}
\bibinfo{author}{Pagani, G.A.}, \bibinfo{author}{Aiello, M.},
  \bibinfo{year}{2013}.
\newblock \bibinfo{title}{The power grid as a complex network: A survey}.
\newblock \bibinfo{journal}{Physica A} \bibinfo{volume}{392},
  \bibinfo{pages}{2688--2700}.
\newblock \DOIprefix\doi{10.1016/j.physa.2013.01.023}.
\bibitem[{Park and Casella(2008)}]{BayesLasso}
\bibinfo{author}{Park, T.}, \bibinfo{author}{Casella, G.},
  \bibinfo{year}{2008}.
\newblock \bibinfo{title}{The {B}ayesian lasso}.
\newblock \bibinfo{journal}{J. Am. Stat. Assoc.} \bibinfo{volume}{103},
  \bibinfo{pages}{681--686}.
\newblock \DOIprefix\doi{10.1198/016214508000000337}.
\bibitem[{Pastor-Satorras et~al.(2015)Pastor-Satorras, Castellano, Van~Mieghem
  and Vespignani}]{Epidemic0}
\bibinfo{author}{Pastor-Satorras, R.}, \bibinfo{author}{Castellano, C.},
  \bibinfo{author}{Van~Mieghem, P.}, \bibinfo{author}{Vespignani, A.},
  \bibinfo{year}{2015}.
\newblock \bibinfo{title}{Epidemic processes in complex networks}.
\newblock \bibinfo{journal}{Rev. Mod. Phys.} \bibinfo{volume}{87},
  \bibinfo{pages}{925--979}.
\newblock \DOIprefix\doi{10.1103/RevModPhys.87.925}.
\bibitem[{Shen et~al.(2014)Shen, Wang, Fan, Di and Lai}]{CS_SIS}
\bibinfo{author}{Shen, Z.}, \bibinfo{author}{Wang, W.X.}, \bibinfo{author}{Fan,
  Y.}, \bibinfo{author}{Di, Z.}, \bibinfo{author}{Lai, Y.C.},
  \bibinfo{year}{2014}.
\newblock \bibinfo{title}{Reconstructing propagation networks with natural
  diversity and identifying hidden sources}.
\newblock \bibinfo{journal}{Nat. Commun.} \bibinfo{volume}{5},
  \bibinfo{pages}{Art. No. 4323}.
\newblock \DOIprefix\doi{10.1038/ncomms5323}.
\bibitem[{Su et~al.(2016)Su, Wang, Wang and Lai}]{CS_geo}
\bibinfo{author}{Su, R.}, \bibinfo{author}{Wang, W.}, \bibinfo{author}{Wang,
  X.}, \bibinfo{author}{Lai, Y.}, \bibinfo{year}{2016}.
\newblock \bibinfo{title}{Data-based reconstruction of complex geospatial
  networks, nodal positioning and detection of hidden nodes}.
\newblock \bibinfo{journal}{R. Soc. Open Sci.} \bibinfo{volume}{3},
  \bibinfo{pages}{Art. No. 150577}.
\newblock \DOIprefix\doi{10.1098/rsos.150577}.
\bibitem[{Tibshirani(1996)}]{lasso_ori}
\bibinfo{author}{Tibshirani, R.}, \bibinfo{year}{1996}.
\newblock \bibinfo{title}{Regression shrinkage and selection via the lasso}.
\newblock \bibinfo{journal}{J. R. Stat. Soc. Ser. B-Stat. Methodol.}
  \bibinfo{volume}{58}, \bibinfo{pages}{267--288}.
\newblock \DOIprefix\doi{https://www.jstor.org/stable/2346178}.
\bibitem[{Tse et~al.(2010)Tse, Liu and Lau}]{TSE2010659}
\bibinfo{author}{Tse, C.K.}, \bibinfo{author}{Liu, J.}, \bibinfo{author}{Lau,
  F.C.}, \bibinfo{year}{2010}.
\newblock \bibinfo{title}{A network perspective of the stock market}.
\newblock \bibinfo{journal}{J. Empir. Financ.} \bibinfo{volume}{17},
  \bibinfo{pages}{659 -- 667}.
\newblock \DOIprefix\doi{10.1016/j.jempfin.2010.04.008}.
\bibitem[{Wainwright and Jordan(2008)}]{VB3}
\bibinfo{author}{Wainwright, M.J.}, \bibinfo{author}{Jordan, M.I.},
  \bibinfo{year}{2008}.
\newblock \bibinfo{title}{Graphical models, exponential families, and
  variational inference}.
\newblock \bibinfo{journal}{Found. Trends Mach. Learn.} \bibinfo{volume}{1},
  \bibinfo{pages}{1--305}.
\newblock \DOIprefix\doi{10.1561/2200000001}.
\bibitem[{{Wang} et~al.(2019){Wang}, {Wang} and {L\"{u}}}]{Biology}
\bibinfo{author}{{Wang}, P.}, \bibinfo{author}{{Wang}, D.},
  \bibinfo{author}{{L\"{u}}, J.}, \bibinfo{year}{2019}.
\newblock \bibinfo{title}{Controllability analysis of a gene network for
  arabidopsis thaliana reveals characteristics of functional gene families}.
\newblock \bibinfo{journal}{IEEE/ACM Trans. Comput. Biol. Bioinform.}
  \bibinfo{volume}{16}, \bibinfo{pages}{912--924}.
\newblock \DOIprefix\doi{10.1109/TCBB.2018.2821145}.
\bibitem[{Wang et~al.(2017)Wang, Yang, Chen, Song, Zhang and Wang}]{Biology3}
\bibinfo{author}{Wang, P.}, \bibinfo{author}{Yang, C.}, \bibinfo{author}{Chen,
  H.}, \bibinfo{author}{Song, C.}, \bibinfo{author}{Zhang, X.},
  \bibinfo{author}{Wang, D.}, \bibinfo{year}{2017}.
\newblock \bibinfo{title}{Transcriptomic basis for drought-resistance in
  {B}rassica napus {L}.}
\newblock \bibinfo{journal}{Sci. Rep.} \bibinfo{volume}{7},
  \bibinfo{pages}{Art. No. 40532}.
\newblock \DOIprefix\doi{10.1038/srep40532}.
\bibitem[{Wang et~al.(2014)Wang, Yu and L\"{u}}]{Biology2}
\bibinfo{author}{Wang, P.}, \bibinfo{author}{Yu, X.}, \bibinfo{author}{L\"{u},
  J.}, \bibinfo{year}{2014}.
\newblock \bibinfo{title}{Identification and evolution of structurally dominant
  nodes in protein-protein interaction networks}.
\newblock \bibinfo{journal}{IEEE Trans. Biomed. Circuits Syst.}
  \bibinfo{volume}{8}, \bibinfo{pages}{87--97}.
\newblock \DOIprefix\doi{10.1109/TBCAS.2014.2303160}.
\bibitem[{Wang et~al.(2016)Wang, Lai and Grebogi}]{recon_review}
\bibinfo{author}{Wang, W.X.}, \bibinfo{author}{Lai, Y.C.},
  \bibinfo{author}{Grebogi, C.}, \bibinfo{year}{2016}.
\newblock \bibinfo{title}{Data based identification and prediction of nonlinear
  and complex dynamical systems}.
\newblock \bibinfo{journal}{Phys. Rep.} \bibinfo{volume}{644},
  \bibinfo{pages}{1--76}.
\newblock \DOIprefix\doi{10.1016/j.physrep.2016.06.004}.
\bibitem[{Wang et~al.(2011a)Wang, Lai, Grebogi and Ye}]{CS_GT}
\bibinfo{author}{Wang, W.X.}, \bibinfo{author}{Lai, Y.C.},
  \bibinfo{author}{Grebogi, C.}, \bibinfo{author}{Ye, J.},
  \bibinfo{year}{2011}a.
\newblock \bibinfo{title}{Network reconstruction based on evolutionary-game
  data via compressive sensing}.
\newblock \bibinfo{journal}{Phys. Rev. X} \bibinfo{volume}{1},
  \bibinfo{pages}{Art. No. 021021}.
\newblock \DOIprefix\doi{10.1103/PhysRevX.1.021021}.
\bibitem[{Wang et~al.(2011b)Wang, Yang, Lai, Kovanis and
  Grebogi}]{CS_Catastrophes}
\bibinfo{author}{Wang, W.X.}, \bibinfo{author}{Yang, R.}, \bibinfo{author}{Lai,
  Y.C.}, \bibinfo{author}{Kovanis, V.}, \bibinfo{author}{Grebogi, C.},
  \bibinfo{year}{2011}b.
\newblock \bibinfo{title}{Predicting catastrophes in nonlinear dynamical
  systems by compressive sensing}.
\newblock \bibinfo{journal}{Phys. Rev. Lett.} \bibinfo{volume}{106},
  \bibinfo{pages}{Art. No. 154101}.
\newblock \DOIprefix\doi{10.1103/PhysRevLett.106.154101}.
\bibitem[{Wang et~al.(2018a)Wang, L\"{u} and Wu}]{CS_tv}
\bibinfo{author}{Wang, X.}, \bibinfo{author}{L\"{u}, J.}, \bibinfo{author}{Wu,
  X.}, \bibinfo{year}{2018}a.
\newblock \bibinfo{title}{Recovering network structures with time-varying nodal
  parameters (in press)}.
\newblock \bibinfo{journal}{IEEE Trans. Syst. Man Cybern. -Syst.}
  \DOIprefix\doi{10.1109/TSMC.2018.2822780}.
\bibitem[{Wang et~al.(2018b)Wang, Meng and Yuan}]{lasso}
\bibinfo{author}{Wang, Y.}, \bibinfo{author}{Meng, D.}, \bibinfo{author}{Yuan,
  M.}, \bibinfo{year}{2018}b.
\newblock \bibinfo{title}{Sparse recovery: from vectors to tensors}.
\newblock \bibinfo{journal}{Natl. Sci. Rev.} \bibinfo{volume}{5},
  \bibinfo{pages}{756--767}.
\newblock \DOIprefix\doi{10.1093/nsr/nwx069}.
\bibitem[{Watts and Strogatz(1998)}]{watts1998collective}
\bibinfo{author}{Watts, D.J.}, \bibinfo{author}{Strogatz, S.H.},
  \bibinfo{year}{1998}.
\newblock \bibinfo{title}{Collective dynamics of ‘small-world’networks}.
\newblock \bibinfo{journal}{Nature} \bibinfo{volume}{393},
  \bibinfo{pages}{440--442}.
\newblock \DOIprefix\doi{10.1038/30918}.
\bibitem[{Wu et~al.(2012)Wu, Wang and Zheng}]{PPGrangerCausality}
\bibinfo{author}{Wu, X.}, \bibinfo{author}{Wang, W.}, \bibinfo{author}{Zheng,
  W.X.}, \bibinfo{year}{2012}.
\newblock \bibinfo{title}{Inferring topologies of complex networks with hidden
  variables.}
\newblock \bibinfo{journal}{Phys. Rev. E} \bibinfo{volume}{86},
  \bibinfo{pages}{046106}.
\newblock \DOIprefix\doi{10.1103/PhysRevE.86.046106}.
\bibitem[{Wu et~al.(2011)Wu, Zhou, Chen and Lu}]{PiecewiseGrangerCausality}
\bibinfo{author}{Wu, X.}, \bibinfo{author}{Zhou, C.}, \bibinfo{author}{Chen,
  G.}, \bibinfo{author}{Lu, J.A.}, \bibinfo{year}{2011}.
\newblock \bibinfo{title}{Detecting the topologies of complex networks with
  stochastic perturbations}.
\newblock \bibinfo{journal}{Chaos} \bibinfo{volume}{21}, \bibinfo{pages}{Art.
  No. 043129}.
\newblock \DOIprefix\doi{10.1063/1.3664396}.
\bibitem[{Xiang et~al.(2018)Xiang, Ma, Chen and Zhang}]{Recon_EM2}
\bibinfo{author}{Xiang, B.B.}, \bibinfo{author}{Ma, C.}, \bibinfo{author}{Chen,
  H.S.}, \bibinfo{author}{Zhang, H.F.}, \bibinfo{year}{2018}.
\newblock \bibinfo{title}{Reconstructing signed networks via ising dynamics}.
\newblock \bibinfo{journal}{Chaos} \bibinfo{volume}{28}, \bibinfo{pages}{Art.
  No. 123117}.
\newblock \DOIprefix\doi{10.1063/1.5053723}.
\bibitem[{Xu and Wang(2017)}]{Epidemic3}
\bibinfo{author}{Xu, S.}, \bibinfo{author}{Wang, P.}, \bibinfo{year}{2017}.
\newblock \bibinfo{title}{Identifying important nodes by adaptive leaderrank}.
\newblock \bibinfo{journal}{Physica A} \bibinfo{volume}{469},
  \bibinfo{pages}{654--664}.
\newblock \DOIprefix\doi{10.1016/j.physa.2016.11.034}.
\bibitem[{Xu et~al.(2017)Xu, Wang and L\"{u}}]{Epidemic2}
\bibinfo{author}{Xu, S.}, \bibinfo{author}{Wang, P.}, \bibinfo{author}{L\"{u},
  J.}, \bibinfo{year}{2017}.
\newblock \bibinfo{title}{Iterative neighbour-information gathering for ranking
  nodes in complex networks}.
\newblock \bibinfo{journal}{Sci. Rep.} \bibinfo{volume}{7},
  \bibinfo{pages}{Art. No. 41321}.
\newblock \DOIprefix\doi{10.1038/srep41321}.
\bibitem[{{Xu} et~al.(2019){Xu}, {Wang}, {Zhang} and {L\"{u}}}]{Epidemic}
\bibinfo{author}{{Xu}, S.}, \bibinfo{author}{{Wang}, P.},
  \bibinfo{author}{{Zhang}, C.}, \bibinfo{author}{{L\"{u}}, J.},
  \bibinfo{year}{2019}.
\newblock \bibinfo{title}{Spectral learning algorithm reveals propagation
  capability of complex networks}.
\newblock \bibinfo{journal}{IEEE Trans. Cybernetics} \bibinfo{volume}{49},
  \bibinfo{pages}{4253--4261}.
\newblock \DOIprefix\doi{10.1109/TCYB.2018.2861568}.
\bibitem[{Zhang et~al.(2019)Zhang, Xu and Zhang}]{VS}
\bibinfo{author}{Zhang, C.X.}, \bibinfo{author}{Xu, S.},
  \bibinfo{author}{Zhang, J.S.}, \bibinfo{year}{2019}.
\newblock \bibinfo{title}{A novel variational {B}ayesian method for variable
  selection in logistic regression models}.
\newblock \bibinfo{journal}{Computat. Stat. Data Analy.} \bibinfo{volume}{133},
  \bibinfo{pages}{1--19}.
\newblock \DOIprefix\doi{10.1016/j.csda.2018.08.025}.
\bibitem[{{Zhang} et~al.(2018){Zhang}, {Yang} and {Wang}}]{CS_weight2}
\bibinfo{author}{{Zhang}, W.}, \bibinfo{author}{{Yang}, G.},
  \bibinfo{author}{{Wang}, L.}, \bibinfo{year}{2018}.
\newblock \bibinfo{title}{Reconstruction of complex time-varying weighted
  networks based on lasso}, in: \bibinfo{booktitle}{2018 37th Chinese Control
  Conference (CCC)}, pp. \bibinfo{pages}{6417--6422}.
\newblock \DOIprefix\doi{10.23919/ChiCC.2018.8484148}.

\end{thebibliography}

\end{document}